\documentclass[10pt,twocolumn,letterpaper]{article}
\pdfoutput=1
\usepackage[cvprfinal]{config/style}
\usepackage[pagebackref=true, breaklinks=true, colorlinks, citecolor=citecolor, bookmarks=false, pagebackref=true]{hyperref}
\usepackage{amsmath}
\usepackage[capitalize]{cleveref}
\usepackage{lipsum}
\usepackage[dvipsnames]{xcolor}
\usepackage{color, colortbl}
\usepackage{amsthm}
\usepackage{wrapfig}
\usepackage{array}
\usepackage{newlfont}
\usepackage{textcomp}
\usepackage{multirow}
\usepackage{commath}
\usepackage{hhline}
\usepackage{tabularx}
\usepackage{bigstrut}
\usepackage{afterpage}
\usepackage{algorithmic}
\usepackage{microtype}
\usepackage{mathtools}
\usepackage{booktabs}
\usepackage[ruled]{algorithm2e}
\usepackage{cite}
\usepackage{dingbat}
\usepackage{balance}

\usepackage{times}
\usepackage{epsfig}
\usepackage[export]{adjustbox}
\usepackage{graphicx}
\usepackage{amssymb}
\usepackage{cuted}
\usepackage{capt-of}
\usepackage{float}
\usepackage{enumitem}
\usepackage{comment}
\usepackage{pifont}
\usepackage[toc,page]{appendix}

\usepackage{soul}
\usepackage{acronym}
\usepackage[usestackEOL]{stackengine}
\usepackage{fancybox}
\usepackage{epigraph}
\usepackage{dirtytalk}
\usepackage{bm}
\usepackage{fontawesome}
\usepackage[accsupp]{axessibility}  % Improves PDF readability for those with disabilities.

\definecolor{citecolor}{HTML}{0071bc}
\definecolor{frontcolor}{HTML}{325ea5}
\definecolor{backcolor}{HTML}{a58b77}
\definecolor{sidecolor}{HTML}{10768c}
\definecolor{skincolor}{HTML}{dcb7b7}
\definecolor{darkred}{rgb}{0.6, 0.1, 0.05}
\definecolor{DeltaColor}{rgb}{0.039,0.73,0.71}
\definecolor{SigmaColor}{rgb}{0.98,0.45,0.0}
\definecolor{AlphaColor}{rgb}{0,0,0.8}
\definecolor{BetaColor}{rgb}{0.8,0,0.8}
\definecolor{GammaColor}{rgb}{0.514,0.34,0.224}
\definecolor{EpsilonColor}{rgb}{0.353,0.725,0.906}
\definecolor{PurpleColor}{HTML}{8B008B}
\definecolor{BadColor}{HTML}{C0392B}
\definecolor{OrangeColor}{rgb}{0.914,0.541,0.0.141}
\definecolor{GreenColor}{rgb}{0.137,0.573,0.565}
\definecolor{RedColor}{rgb}{0.949,0.275, 0.224}
\definecolor{LightCyan}{rgb}{0.88,1,1}
\definecolor{Gray}{gray}{0.85}

\definecolor{bestcolor}{rgb}{1, 0.5, 0.25}
\definecolor{secondbestcolor}{rgb}{1, 0.8, 0.5}
\newcommand{\bone}{\cellcolor{bestcolor}}
\newcommand{\btwo}{\cellcolor{secondbestcolor}}

\newcolumntype{a}{>{\columncolor{Gray}}c}

\newcommand{\change}[1]{{\color{Black} #1}}
\newcommand{\camera}[1]{{\color{Black} #1}}
\newcommand{\suppl}{\camera{SupMat.}\xspace}
\newcommand{\video}{\href{https://econ.is.tue.mpg.de}{video on our website}\xspace}
\newcommand{\specific}[1]{\xspace{\text{\fontfamily{qcr}\selectfont #1}}\xspace}

\newcommand{\cheading}[1]{\noindent\textbf{#1.}}
\newcommand{\qheading}[1]{\noindent\textbf{#1.}}

\newcommand{\zheading}[1]{\textbf{#1.}}

\newcommand{\TODO}[1]{\textcolor{black}{#1}\xspace}

\newcommand{\SUBSPRINT}[1]{\textcolor{black}{#1}\xspace}

\newcommand{\xmark}{\textcolor{RedColor}{\ding{55}}\xspace}
\newcommand{\cmark}{\textcolor{GreenColor}{\ding{51}}\xspace}

\newcommand{\colorRef}[1]{\textcolor{red}{#1}}
\crefname{figure}{\colorRef{Fig.}}{\colorRef{Figs.}}
\Crefname{figure}{\colorRef{Figure}}{\colorRef{Figures}}
\crefname{section}{\colorRef{Sec.}}{\colorRef{Secs.}}
\Crefname{section}{\colorRef{Section}}{\colorRef{Sections}}
\crefname{table}{\colorRef{Tab.}}{\colorRef{Tabs.}}
\Crefname{table}{\colorRef{Table}}{\colorRef{Tables}}
\Crefname{equation}{\colorRef{Eq.}}{\colorRef{Eqs.}}
\Crefname{equation}{\colorRef{Equation}}{\colorRef{Equation}}

\renewcommand{\etc}{\mbox{etc}\xspace}
\renewcommand{\etal}{\mbox{et al.}\xspace}
\renewcommand{\ie}{\mbox{i.e.}\xspace}
\renewcommand{\eg}{\mbox{e.g.}\xspace}

\newcolumntype{x}[1]{>{\centering\arraybackslash}p{#1pt}}
\newcolumntype{y}[1]{>{\raggedright\arraybackslash}p{#1pt}}
\newcolumntype{z}[1]{>{\raggedleft\arraybackslash}p{#1pt}}
\newlength\savewidth\newcommand\shline{\noalign{\global\savewidth\arrayrulewidth
  \global\arrayrulewidth 1pt}\hline\noalign{\global\arrayrulewidth\savewidth}}

\newcommand{\rgb}{\mbox{RGB}\xspace}

\newcommand{\smpl}{\mbox{SMPL}\xspace}
\newcommand{\smplx}{\mbox{SMPL-X}\xspace}
\newcommand{\smplX}{\smplx}

\newcommand{\groundtruth}{{ground-truth}\xspace}

\newcommand{\sota}{\mbox{SOTA}\xspace}

\newcommand{\inthewild}{\mbox{in-the-wild}\xspace}
\newcommand{\itw}{\inthewild}
\newcommand{\ood}{{out-of-distribution}\xspace}
\newcommand{\OOD}{\mbox{OOD}\xspace}

\newcommand{\twoD}{2D\xspace}
\newcommand{\threeD}{3D\xspace}
\newcommand{\twoFiveD}{\mbox{2.5D}\xspace}

\newcommand{\pymafx}{\mbox{PyMAF-X}\xspace}

\newcommand{\mcubes}{\mbox{Marching cubes}\xspace}

\newcommand{\pixie}{\mbox{PIXIE}\xspace}

\newcommand{\pifu}{\mbox{PIFu}\xspace}
\newcommand{\pifuHD}{\mbox{PIFuHD}\xspace}
\newcommand{\pifuorHD}{\mbox{PIFu(HD)}\xspace}

\newcommand{\pina}{\mbox{PINA}\xspace}

\newcommand{\selfportrait}{\mbox{Self-Portraits}\xspace}
\newcommand{\jiff}{\mbox{JIFF}\xspace}
\newcommand{\smplicit}{\mbox{SMPLicit}\xspace}

\newcommand{\clothwild}{\mbox{ClothWild}\xspace}

\newcommand{\pifuhd}{\mbox{PIFuHD}\xspace}
\newcommand{\geopifu}{\mbox{GeoPIFu}\xspace}
\newcommand{\pamir}{\mbox{PaMIR}\xspace}
\newcommand{\arch}{\mbox{ARCH}\xspace}
\newcommand{\archplus}{\mbox{ARCH++}\xspace}

\newcommand{\ifnet}{\mbox{IF-Nets}\xspace}
\newcommand{\ifnetplus}{\mbox{IF-Nets+}\xspace}

\newcommand{\icon}{\mbox{ICON}\xspace}

\newcommand{\facsimile}{\mbox{FACSIMILE}\xspace}
\newcommand{\modulehuman}{\mbox{Moduling Humans}\xspace}

\newcommand{\maskrcnn}{\mbox{Mask R-CNN}\xspace}

\newcommand{\renderppl}{\mbox{Renderpeople}\xspace}
\newcommand{\thuman}{\mbox{THuman2.0}\xspace}

\newcommand{\cape}{\mbox{CAPE}\xspace}

\newcommand{\capeNFP}{\mbox{``CAPE-NFP''}\xspace}

\newcommand{\modelname}{\mbox{ECON}\xspace} 
\newcommand{\modelnameLong}{Explicit Clothed humans Optimized via Normal integration\xspace}
\newcommand{\ourtitle}{\camera{\modelname: \modelnameLong}\xspace}

\newcommand{\bni}{\mbox{BiNI}\xspace}
\newcommand{\dbni}{\mbox{d-BiNI}\xspace}
\newcommand{\dBNI}{\dbni}

\newcommand{\IF}{\mbox{IF}\xspace}

\newcommand{\projectURL}{\href{https://econ.is.tue.mpg.de}{\tt{econ.is.tue.mpg.de}}}

\DeclareMathAlphabet\mathbfcal{OMS}{cmsy}{b}{n}
\newcommand{\bodyMesh}{\mathcal{M}^\text{b}}
\newcommand{\cullbodyMesh}{\mathcal{M}^\text{cull}}

\newcommand{\FrontSurface}{\mathcal{M_\text{F}}}
\newcommand{\BackSurface}{\mathcal{M_\text{B}}}

\newcommand{\BniSurfaces}{\mathcal{M_\text{\{F,B\}}}}
\newcommand{\BniDepths}{\widehat{\mathcal{Z}}^\text{c}_\text{\{F,B\}}}

\newcommand{\inputimg}{\mathcal{I}}

\newcommand{\bodyDepthImg}{\mathcal{Z}^{\text{b}}}
\newcommand{\bodyDepthImgF}{\mathcal{Z}^{\text{b}}_{\text{F}}}
\newcommand{\bodyDepthImgB}{\mathcal{Z}^{\text{b}}_{\text{B}}}

\newcommand{\clothDepthImg}{\widehat{\mathcal{Z}}^{\text{c}}}
\newcommand{\clothDepthImgF}{\ensuremath{\widehat{\mathcal{Z}}^{\text{c}}_{\text{F}}}}
\newcommand{\clothDepthImgB}{\ensuremath{\widehat{\mathcal{Z}}^{\text{c}}_{\text{B}}}}

\newcommand{\gtclothDepthImgF}{\ensuremath{\mathcal{Z}^{\text{c}}_{\text{F}}}}
\newcommand{\gtclothDepthImgB}{\ensuremath{\mathcal{Z}^{\text{c}}_{\text{B}}}}

\newcommand{\bodyNormImg}{\mathcal{N}^{\text{b}}}

\newcommand{\normG}{\mathcal{G}^{\text{N}}}

\newcommand{\predCloNormImg}{\widehat{\mathcal{N}}^{\text{c}}}
\newcommand{\predCloNormImgF}{\ensuremath{\widehat{\mathcal{N}}^{\text{c}}_\text{F}}}
\newcommand{\predCloNormImgB}{\ensuremath{\widehat{\mathcal{N}}^{\text{c}}_\text{B}}}

\newcommand{\bniLoss}{\ensuremath{\mathcal{L}_{\text{n}}}\xspace}
\newcommand{\depthPriorLoss}{\ensuremath{\mathcal{L}_{\text{d}}}\xspace}
\newcommand{\silhouetteLoss}{\ensuremath{\mathcal{L}_{\text{s}}}\xspace}

\newcommand{\depthFront}{\ensuremath{\widehat{\mathbf{z}_{\text{F}}}} \xspace}
\newcommand{\depthBack}{\ensuremath{\widehat{\mathbf{z}_{\text{B}}}} \xspace}

\newcommand{\depthFrontPrior}{\ensuremath{\mathbf{z}_\text{F}} \xspace}
\newcommand{\depthBackPrior}{\ensuremath{\mathbf{z}_\text{B}} \xspace}
\newcommand{\depthCombined}{\ensuremath{\widehat{\mathbf{z}}} \xspace}
\newcommand{\depthCombinedPrior}{\ensuremath{\mathbf{z}} \xspace}
\newcommand{\domainN}{\ensuremath{\Omega_{\text{n}}}\xspace}
\newcommand{\domainD}{\ensuremath{\Omega_{\text{z}}}\xspace}
\newcommand{\silhouette}{\ensuremath{\partial\Omega_{\text{n}}}\xspace}
\newcommand{\V}[1]{\mathbf{#1}}

\newcommand{\gtCloNormImgFB}{\mathcal{N}^{\text{c}}}

\newcommand{\body}{\text{b}}
\newcommand{\cloth}{\text{c}}
\newcommand{\normal}{\text{N}}
\newcommand{\joint}{\text{J}}
\newcommand{\mask}{\text{S}}
\newcommand{\recMesh}{\mathcal{R}}
\newcommand{\recMeshif}{\mathcal{R}_\text{IF}}

\acrodef{amt}[AMT]{Amazon Mechanical Turk}

\newcommand{\myparagraph}[1]{\vspace{0.075in}\noindent\textbf{#1}}

\makeatletter
\newcommand*{\addFileDependency}[1]{% argument=file name and extension
  \typeout{(#1)}
  \@addtofilelist{#1}
  \IfFileExists{#1}{}{\typeout{No file #1.}}
}
\makeatother

\def\cvprPaperID{443}
\def\confName{CVPR}
\def\confYear{2023}

\begin{document}

\title{\vspace{-0.20em}\ourtitle\vspace{-0.20em}}

\author{
Yuliang Xiu$^1$ \quad Jinlong Yang$^1$ \quad Xu Cao$^2$ \quad Dimitrios Tzionas$^3$ \quad Michael J. Black$^1$\\
{\normalsize $^1$Max Planck Institute for Intelligent Systems, T{\"u}bingen, Germany}\\
{\normalsize $^2$Osaka University, Japan \quad $^3$University of Amsterdam, the Netherlands}\\
{\tt\small \{yuliang.xiu, jinlong.yang, black\}@tue.mpg.de \quad cao.xu@ist.osaka-u.ac.jp \quad d.tzionas@uva.nl}\\
}

\newcommand{\styleFront}{\textbf{\textcolor{frontcolor}{front}}\xspace}
\newcommand{\styleBack}{\textbf{\textcolor{backcolor}{back}}\xspace}
\newcommand{\styleMissingGeometry}{\textbf{\textcolor{sidecolor}{missing geometry}}\xspace}
\newcommand{\styleFaceAndHands}{\textbf{\textcolor{skincolor}{Face or hands}}\xspace}

\newcommand{\teaserCaption}{
\cheading{Human digitization from a color image}
% \modelname combines the best aspects of implicit and explicit surfaces to infer high-fidelity \threeD humans, even with loose clothing or in challenging poses.
\camera{\modelname combines the best aspects of free-form implicit representation, and explicit anthropomorphic regularization to infer high-fidelity \threeD humans, even with loose clothing or in challenging poses.}
It does so in three steps: 
(1) It infers detailed \twoD normal maps for the front and back side (\cref{sec: detailed normal prediction}).
(2) The normal maps are converted into detailed, yet incomplete, \twoFiveD 
\styleFront and 
\styleBack surfaces 
guided by a \smplx estimate (\cref{sec: front and back surface reconstruction}). 
(3) It then ``inpaints'' the \styleMissingGeometry between two surfaces 
(\cref{sec: human shape completion}). 
\camera{\styleFaceAndHands can be optionally replaced with the cleaner ones from \smplX.} 
See the \video for more results.
}

\twocolumn[{
    \renewcommand\twocolumn[1][]{#1}
    \maketitle
    \centering
    \vspace{-0.5em}
    \begin{minipage}{1.00\textwidth}
        \centering
        \includegraphics[trim=000mm 000mm 000mm 000mm, clip=False, width=\linewidth]{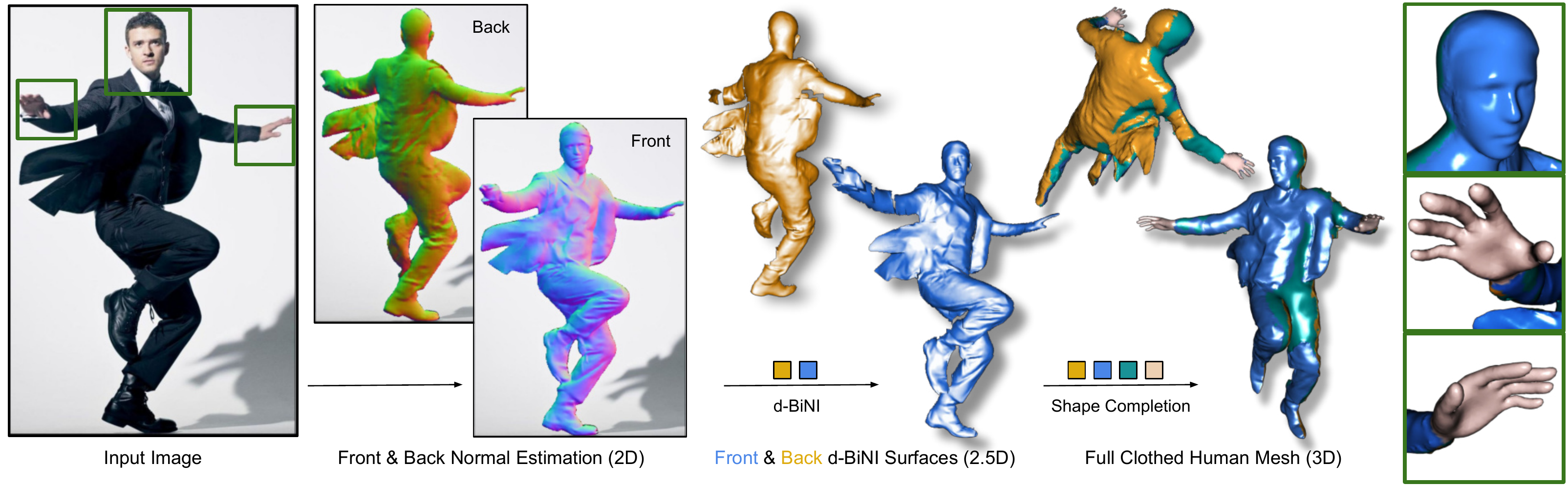}
    \end{minipage}
    \vspace{-0.5 em}
    \captionsetup{type=figure}
    \captionof{figure}{\teaserCaption}
    \label{fig:teaser}
    \vspace{1.2em}
}]

\begin{abstract}
\vspace{-1.0 em}
The combination of deep learning, artist-curated scans, and Implicit Functions (\IF), is enabling the creation of detailed, clothed, \threeD humans from images.
However, existing methods are far from perfect. 
\IF-based methods recover free-form geometry, but produce disembodied limbs or degenerate shapes for novel poses or clothes. 
To increase robustness for these cases, existing work uses an explicit parametric body model to constrain surface reconstruction, but this limits the recovery of free-form surfaces such as loose clothing that deviates from the body.
\camera{What we want is a method that combines the best properties of implicit representation and explicit body regularization.}
To this end, we make two key observations:
(1) current networks are better at inferring detailed \twoD maps than full-\threeD surfaces, and 
(2) a parametric model can be seen as a ``canvas'' for stitching together detailed surface patches. 
Based on these, our method, \modelname, has three main steps:
(1) It infers detailed \twoD normal maps for the front and back side of a clothed person. 
(2) From these, it recovers \twoFiveD front and back surfaces, called \dBNI, that are equally detailed, yet incomplete, and registers these \wrt each other with the help of a \smplX body mesh recovered from the image. 
(3) It ``inpaints'' the missing geometry between \dBNI surfaces. 
\SUBSPRINT{If the face and hands are noisy, they can optionally be replaced with the ones of \smplX.}
As a result, \modelname infers high-fidelity \threeD humans even in loose clothes and challenging poses. 
% \camera{Quantitative evaluation on the \cape and \renderppl datasets reveals that \modelname is more accurate than previous methods.}
\camera{This goes beyond previous methods, according to the quantitative evaluation on the \cape and \renderppl datasets}.
%
%
% Quantitative evaluation on 
% the \cape and \renderppl datasets 
% shows that \modelname is more accurate than the state of the art. 
Perceptual studies also show that \modelname's perceived realism is 
better \TODO{by a large margin}. 
Code and models are available for research purposes at \projectURL 
\end{abstract}

\section{Introduction}
\label{sec: introduction}

Human avatars will be key for future games and movies, mixed-reality, tele-presence and the ``metaverse''. 
To build realistic and personalized avatars at scale, 
we need 
to faithfully reconstruct detailed \threeD humans from color 
photos taken in the wild.
This is still an open problem, due to its challenges; 
people wear all kinds of different
clothing
and accessories, 
and they pose their bodies in many, often imaginative, ways. 
A good reconstruction method must accurately capture these, while also 
being robust to novel clothing and poses. 

\begin{figure}[t]
	\centering      %      l     b     r     u
	\vspace{-0.7 em}
	\includegraphics[trim=000mm 000mm 000mm 000mm, clip=true, width=\linewidth]{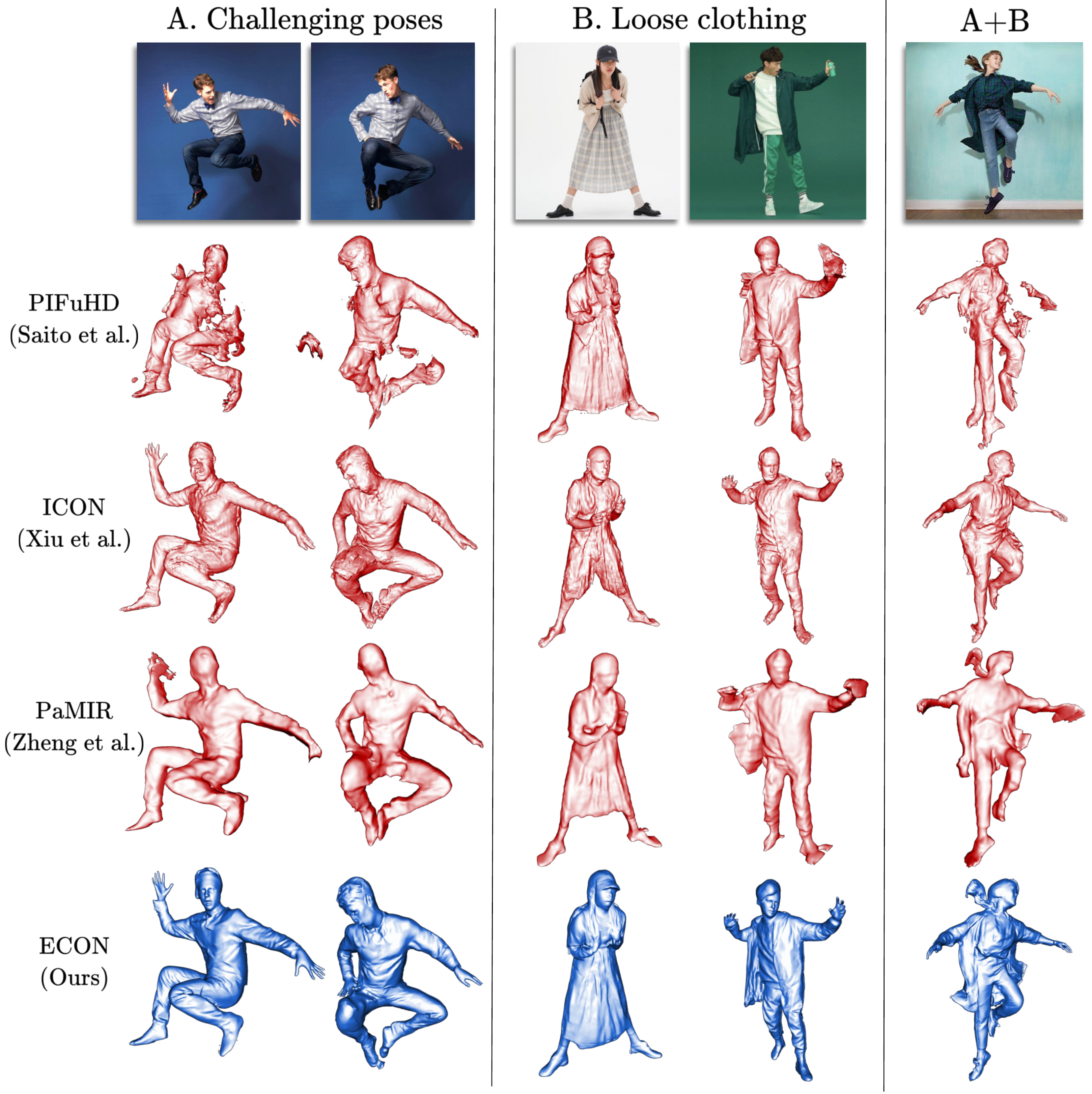}
	\vspace{-2.2 em}
	\caption{
                \cheading{Summary of \sota}
	            \textcolor{BadColor}{\pifuHD}~\cite{saito2020pifuhd} recovers clothing details, but struggles with novel poses. 
	            \textcolor{BadColor}{\icon}~\cite{xiu2022icon} and \textcolor{BadColor}{\pamir}~\cite{zheng2020pamir} regularize shape to a body shape, but over-constrain the skirts, or over-smooth the wrinkles. 
	            \textcolor{frontcolor}{\modelname} combines their best aspects.
    }
	\vspace{-1.5 em}
	\label{fig:motivation}
\end{figure}

Initial, promising, results have been made possible by using
artist-curated scans as training data, and \change{implicit functions} (\IF)~\cite{park2019deepSDF, mescheder2019occNet} as the \threeD representation. 
Seminal work 
on \pifuorHD~\cite{saito2019pifu,saito2020pifuhd} uses ``pixel-aligned'' \IF and reconstructs clothed \threeD humans with unconstrained topology. 
However, these methods tend to overfit 
to the poses seen in the training data, and have no explicit
knowledge about the human body's structure. 
Consequently, they produce disembodied limbs or degenerate shapes for images with novel poses; 
see the 2nd row of \cref{fig:motivation}. 
Follow-up work 
\cite{he2020geoPifu, zheng2020pamir, xiu2022icon} 
accounts for such artifacts by 
regularizing the \IF using a shape prior provided by an explicit body model~\cite{SMPL:2015, pavlakos2019expressive}, 
\camera{but regularization introduces a topological constraint, restricting generalization to novel clothing} %styles 
while attenuating shape details; 
see the 3rd and 4th rows of \cref{fig:motivation}. 
In a nutshell, there are trade-offs between robustness, generalization and detail.

What we want is
the \emph{best of both worlds}; that is, the robustness of explicit anthropomorphic body models, and the flexibility of \IF to capture arbitrary clothing topology. 
To that end, 
we make two key observations: 
% (1) \change{While it has been shown that \twoD normal maps can be effortlessly predicted from color image~\cite{Jafarian_2021_CVPR_TikTok, xiu2022icon, saito2020pifuhd}, 
(1) \change{While inferring detailed \twoD normal maps from color images is relatively easy~\cite{Jafarian_2021_CVPR_TikTok, xiu2022icon, saito2020pifuhd}, 
inferring \threeD geometry with equally fine details is still challenging~\cite{chen2022gdna}}.
Thus, we exploit networks to infer detailed ``geometry-aware'' \twoD maps that we then lift to \threeD.
(2) A body model can be seen as a low-frequency ``canvas'' that ``guides'' the stitching of detailed surface parts. 

With these in mind, we develop \modelname, which stands for ``\modelnameLong''. 
It takes,  as input,
an \rgb image  and a \smplX body inferred from the image.
Then, it outputs a \threeD human in free-form clothing %and body pose, 
with a level of detail and robustness that goes beyond the state of the art (\sota); 
see the bottom of~\cref{fig:motivation}. 
Specifically, \modelname has \emph{three steps}. 

\zheading{Step 1: Front \& back normal reconstruction}
We predict front- and back-side clothed-human normal maps from the input \rgb image, conditioned on the body estimate, with a standard image-to-image translation network. 
 
\zheading{Step 2: Front \& back surface reconstruction}
We take the previously predicted normal maps, 
and the corresponding 
depth maps that are rendered from the \smplx mesh,  
to produce detailed and coherent front-/back-side \threeD surfaces, $\{\FrontSurface, \BackSurface\}$. 
To this end, we extend the recent \bni method \cite{xu2022bilateral}, and develop a novel optimization scheme that is aimed at satisfying three goals for the resulting surfaces:
(1) 
their high-frequency components agree with clothed-human normals, 
(2) 
their low-frequency components and the discontinuities agree with the \smplX ones, and 
(3) 
the depth values on their silhouettes are coherent with each other and consistent with the 
\smplX-based depth maps. 
The two output surfaces, $\{\FrontSurface, \BackSurface\}$, 
are detailed yet incomplete, \ie, there is missing geometry
in occluded and ``profile'' regions.

\zheading{Step 3: Full
\threeD shape
completion}
This module takes two inputs:
(1) the \smplX mesh, and 
(2) the 
two 
\dBNI surfaces, $\{\FrontSurface, \BackSurface\}$.
The goal is to ``inpaint'' the missing geometry.
Existing methods struggle with this problem. 
On one hand, Poisson reconstruction~\cite{kazhdan2013screened} produces ``blobby'' shapes 
and naively ``infills'' holes without exploiting a shape distribution prior.
On the other hand,  
data-driven approaches, such as \ifnet~\cite{chibane20ifnet},
struggle with missing parts caused by (self-)occlusions,
% and lose details present in given high-quality surfaces, 
\camera{and fail to keep the fine details present on two \dBNI surfaces,}
producing degenerate geometries.

We address above the limitations in two steps:
(1)
We extend and re-train \ifnet to be conditioned on the \smplX body, so that \smplX regularizes shape ``infilling''. 
We discard the triangles that lie close to $\{\FrontSurface, \BackSurface\}$, and keep the remaining ones as ``infilling patches''. 
(2)
We stitch together the front- and back-side surfaces and infilling patches via Poisson reconstruction; note that holes between these are small enough for a general purpose method. 
The result is a full \threeD shape of a clothed human; see  \cref{fig:motivation}, bottom.

We evaluate \modelname both on established benchmarks 
(\cape \cite{ma2020cape} and \renderppl \cite{renderpeople}) 
and \itw images. 
Quantitative analysis reveals 
% that \modelname outperforms the \sota
\modelname's superiority. A perceptual study echos this, showing that \modelname is significantly preferred over competitors on challenging poses and loose clothing, and competitive with \pifuhd on fashion images. Qualitative results show that \modelname generalizes better than the \sota to a wide variety of poses and clothing, even with extreme looseness or complex topology;
see \cref{fig: qualitative figure}. 

% \modelname combines the best aspects of explicit shape regularization and free-from implicit representation, % modeling, 
% to recover \threeD clothed humans %from a color image 
% with a 
% good 
% level of detail and robustness.

\camera{With both pose-robustness and topological flexibility, \modelname recovers \threeD clothed humans 
with a 
good 
level of detail and realistic pose.}
Code and models are available for research purposes at \projectURL

\section{Related Work}

\begin{figure*}
    \centerline{
    \includegraphics[trim=000mm 000mm 000mm 000mm, clip=True, width=1.0\linewidth]{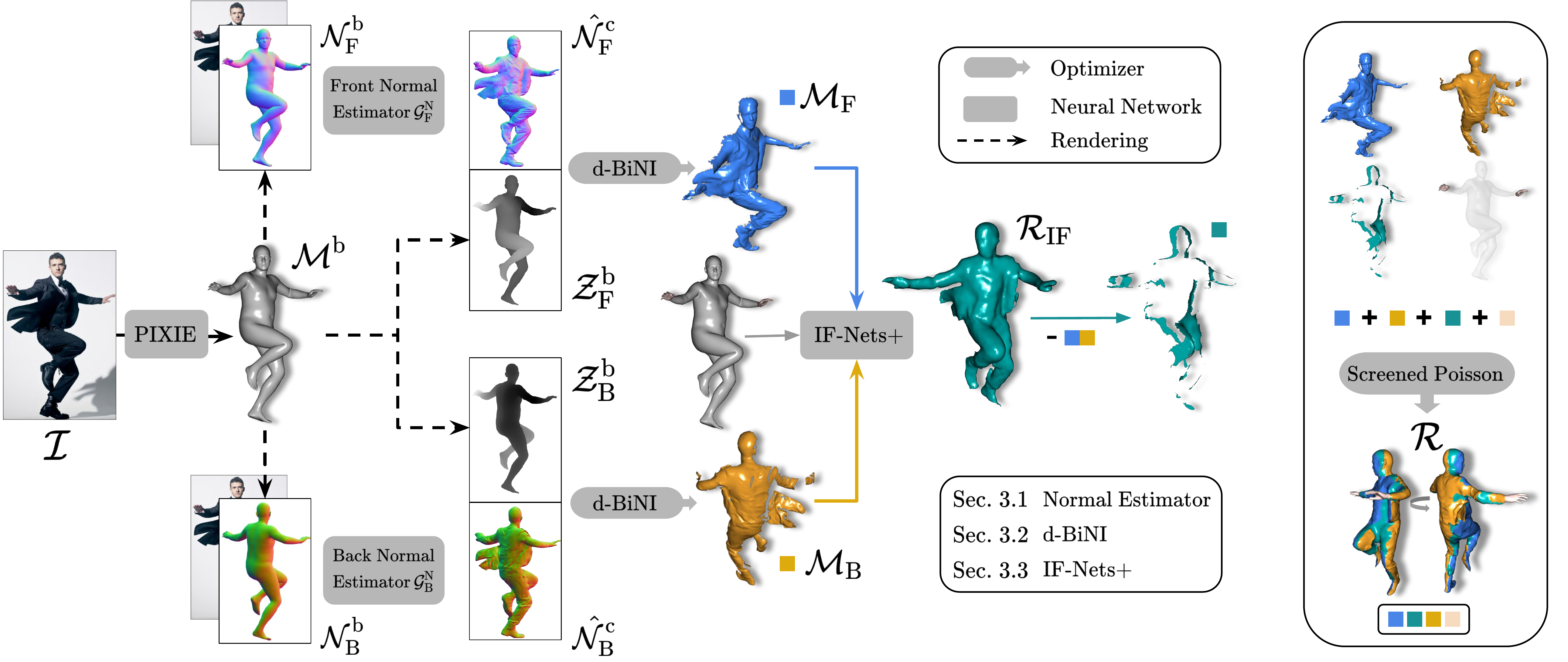}}
    % \vspace{-0.1in}
    \caption{
                \cheading{Overview}
                \modelname takes as input an \rgb image, $\bm{\inputimg}$, and a \smplx body, $\bm{\bodyMesh}$. 
                Conditioned on the rendered front and back body normal images, $\bm{\bodyNormImg}$, \modelname first predicts front and back clothing normal maps, $\bm{\predCloNormImg}$. 
                These two maps, along with body depth maps, $\bm{\bodyDepthImg}$, are fed into \change{a} \dbni optimizer to produce front and back surfaces, $\{\bm{\textcolor{frontcolor}{\FrontSurface},\textcolor{backcolor}{\BackSurface}}\}$. 
                Based on such partial surfaces, and body estimate $\bm{\bodyMesh}$, \ifnetplus implicitly completes $\bm{\textcolor{sidecolor}{\recMeshif}}$. With optional \styleFaceAndHands from $\bm{\bodyMesh}$, screened Poisson combines everything as final watertight $\bm{\recMesh}$.
    }
    \label{fig:architecture}
    \vspace{-0.1in}
\end{figure*}

\paragraph{Image-based clothed human reconstruction.}
Regarding geometric representation, we group the mainstream clothed human reconstruction approaches into ``implicit'' and ``explicit''. \camera{Note that with the terms implicit/explicit we mainly refer 
to the \textit{surface decoder} rather than the \textit{feature encoder}.}

\textbf{1) Explicit-shape-based approaches} use either a mesh-based parametric body model~\cite{Joo2018_adam,SMPL:2015,pavlakos2019expressive,romero2017embodied,xu2020ghum},  or a non-parametric
depth map~\cite{gabeur2019moulding,smith2019facsimile} or point cloud~\cite{zakharkin2021point}, to reconstruct \threeD humans. 
Many methods \cite{kanazawa2018hmr,kolotouros2019spin,sun2021BEV,li2021hybrik,zhang2021pymaf,yao2021pixie,pymafx2022,Kocabas2021pare,yi2022generating,tripathi2023ipman,li2023niki} estimate or regress minimally-clothed \threeD body meshes from \rgb pixels and ignore clothing. 
To account for clothed human shapes, another line of work \cite{alldieck2018videoavatar,alldieck2019peopleInClothing,zhu2019hierarchMeshDeform,lazova2019textures360,ma2020cape,ponsMoll2017clothCap,alldieck2019tex2shape,xiang2020monoClothCap} adds \threeD offsets on top of the body mesh. 
This is compatible with current animation pipelines, as they inherit the hierarchical skeleton and skinning weights from the underling statistical body model. 
However, this ``body+offset'' approach is not flexible enough to model loose clothing, which deviates significantly from the body topology, such as dresses and skirts. 
To  increase  topological flexibility, some methods \cite{bhatnagar2019multiGarmentNet,jiang2020bcnet} reconstruct  \threeD clothed humans by  identifying the type of clothing and using the appropriate model to reconstruct it.
Scaling up this ``cloth-aware'' approach to many clothing styles is nontrivial, limiting generalization to \inthewild outfit variation.

\textbf{2) Implicit-function-based approaches} are topology-agnostic and, thus, can be used to represent arbitrary 3D clothed human shapes. 
\smplicit~\cite{corona2021smplicit}, \clothwild~\cite{Moon_2022_ECCV_ClothWild} and DIG~\cite{li2022dig} learn a generative clothing model with neural distance fields~\cite{park2019deepSDF, mescheder2019occNet, chibane2020ndf} from a \threeD clothing dataset. 
Given an image, the clothed human is reconstructed by estimating a parametric body and optimizing the latent space of the clothing model. However, the results usually do not align well with the image and lack %of 
geometric detail.

\pifu~\cite{saito2019pifu} introduces pixel-aligned implicit human shape reconstruction and \pifuhd~\cite{saito2020pifuhd} significantly improves the geometric details with a multi-level architecture and normal maps predicted from the \rgb image.
However, these two methods do not exploit knowledge of the human body structure. 
Therefore, these methods overfit to the body poses in the training data, \eg fashion poses. 
They fail to generalize to novel poses, producing non-human shapes with broken or disembodied limbs. 
To address these issues, several methods introduce different geometric priors to regularize the deep implicit representation: \geopifu~\cite{he2020geoPifu} introduces a coarse shape of volumetric humans, \selfportrait~\cite{Li2020portrait}, \pina~\cite{dong2022pina}, and S3~\cite{yang2021s3} use depth or LIDAR information to regularize shape and improve robustness to pose variation. 

Another direction leverages parametric body models, which represent human body shape well, model the kinematic structure of the body, and can be reliably estimated from  \rgb images of clothed people.
Such a representation can be viewed as a base shape upon which to model clothed humans.
Therefore, several methods combine parametric body models with expressive implicit representations to get the best of both worlds. 
\pamir~\cite{zheng2020pamir} and DeepMultiCap~\cite{zheng2021deepmulticap} condition the pixel-aligned features on a posed and voxelized \smpl mesh. \jiff Introduces a 3DMM face prior to improve the realism of the facial region. \arch~\cite{huang2020arch}, \archplus~\cite{he2021ICCVarchplus} and CAR~\cite{liao2023car} use \smpl to unpose the pixel-aligned query points from a posed space to a canonical space. To further generalize to unseen poses on \itw photos, \icon~\cite{xiu2022icon} regresses shapes from locally-queried features.
However, the above approaches gain robustness to unseen poses at the cost of %their 
generalization ability to various, especially loose, clothing topologies. 
We argue that this is because loose clothing differs greatly from human body and that conditioning on the \smpl body in 3D makes it harder for networks to make full use of 2D image features. % and 3D body information.

Our work is also inspired by ``sandwich-like'' monocular reconstruction approaches, represented by \modulehuman~\cite{gabeur2019moulding}, \facsimile~\cite{smith2019facsimile} and Any-Shot GIN~\cite{Xian2022gin}. \modulehuman has two networks: a \textit{generator} that estimates the visible (front) and invisible (back) depth maps from \rgb images, and a \textit{discriminator} that helps regularize the estimation via an adversarial loss. 
\facsimile further improves the geometric details by leveraging a normal loss, which is directly computed from depth estimates via differentiable layers. 
Recently, Any-Shot GIN generalizes the sandwich-like scheme to novel classes of objects. 
Given \rgb images, it predicts front and back depth maps as well, and then exploits \ifnet~\cite{chibane20ifnet} for shape completion. 
We follow a similar path and extend it, to successfully reconstruct clothed human shapes with \sota pose generalization, and better details from normal images.

\section{Method}
\label{sec: method}

Given an \rgb image, \modelname first estimates front and back normal maps (\cref{sec: detailed normal prediction}), then converts them into front and back partial surfaces (\cref{sec: front and back surface reconstruction}), and finally ``inpaints'' the missing geometry with the help of \ifnetplus (\cref{sec: human shape completion}). See \modelname's overview in~\cref{fig:architecture}.

\subsection{Detailed normal map prediction}
\label{sec: detailed normal prediction}

Trained on abundant pairs of \rgb images and normal images, a 
``front" 
normal map, $\predCloNormImgF$, can be accurately estimated from an \rgb image using image-to-image translation networks, as demonstrated in \pifuhd~\cite{saito2020pifuhd} or \icon~\cite{xiu2022icon}. 
Both methods \SUBSPRINT{also} infer a %the 
``back'' 
normal map, $\predCloNormImgB$, from the image.
But, the absence of image cues leads to over-smooth $\predCloNormImgB$. 
To address this, we fine-tune \icon's backside normal \camera{predictor, $\normG_\text{B}$}, with an additional MRF loss~\cite{wang2018image} \camera{to enhance the local details by minimizing the difference between the predicted $\predCloNormImg$ and ground truth (GT) $\gtCloNormImgFB$ in feature space.}

To guide the normal map prediction and make it robust to various body poses, \icon conditions the normal map prediction module on the body normal maps, $\bodyNormImg$, rendered from the estimated body $\bodyMesh$. 
Thus, it is important to accurately align the estimated body and clothing silhouette. Apart from the $\mathcal{L}_{\normal\text{\_diff}}$ and $\mathcal{L}_{\text{S}\text{\_diff}}$ used in \icon~\cite{xiu2022icon}, we also apply \twoD body landmarks in an additional loss term, $\mathcal{L}_{\text{J}\text{\_diff}}$, to further optimize the \smplx body, $\bodyMesh$, inferred from \pixie~\cite{yao2021pixie} or \pymafx~\cite{pymafx2022}. 
Specifically, we optimize \smplx's shape, $\beta$, pose, $\theta$, and translation, $t$, to minimize:
\begin{equation}
\begin{gathered}
    \label{eq:body-fit}
    \mathcal{L}_\text{\smplx} = \mathcal{L}_{\normal\text{\_diff}} + \mathcal{L}_{\mask\text{\_diff}} + \mathcal{L}_{\joint\text{\_diff}} \text{,}\\
    \mathcal{L}_{\joint\text{\_diff}} = \lambda_\text{\joint\text{\_diff}}| \mathcal{J}^{\body}-\widehat{\mathcal{J}^{\cloth}} | ,
\end{gathered}
\end{equation}
where $\mathcal{L}_{\normal\text{\_diff}}$ and $\mathcal{L}_{\mask\text{\_diff}}$ are the normal-map loss and silhouette loss introduced in \icon~\cite{xiu2022icon}, and $\mathcal{L}_{\joint\text{\_diff}}$ is the joint loss (L2) between \twoD landmarks $\widehat{\mathcal{J}^{\cloth}}$, \camera{which are estimated by a
% \offtheshell 
\twoD keypoint estimator} from the \rgb image $\inputimg$, and the corresponding re-projected \twoD joints $\mathcal{J}^{\body}$ from $\bodyMesh$. 
\change{For more implementation details, see \cref{sec: suppl-normal-prediction} in~\suppl}

\subsection{Front and back surface reconstruction}
\label{sec: front and back surface reconstruction}

We now lift the clothed normal maps to \twoFiveD surfaces.
We expect these \twoFiveD surfaces to satisfy three conditions: 
(1) high-frequency surface details agree %are consistent 
with predicted clothed normal maps, 
(2) low-frequency surface variations, including discontinuities, 
\SUBSPRINT{agree with \smplx's ones}, 
and 
(3) 
\SUBSPRINT{the depth of the front and back silhouettes} 
are close to each other. 

Unlike \pifuhd~\cite{saito2020pifuhd} or \icon~\cite{xiu2022icon}, which train a neural network to regress the implicit surface from normal maps, we explicitly model the depth-normal relationship using variational normal integration methods~\cite{queau2018normal,xu2022bilateral}.
Specifically, we tailor the recent bilateral normal integration (\bni) method~\cite{xu2022bilateral} to full-body mesh reconstruction by harnessing the coarse prior, depth maps, and silhouette consistency.

To satisfy the three conditions, we propose a depth-aware silhouette-consistent bilateral normal integration (\dbni) method to jointly optimize for the front and back clothed depth maps, $\clothDepthImgF$ and~$\clothDepthImgB$:
\begin{align}
    \dbni(\predCloNormImgF, \predCloNormImgB, \bodyDepthImgF, \bodyDepthImgB) \rightarrow{\clothDepthImgF, \clothDepthImgB}.
    \label{eq.dbni_full}
\end{align}
Here, $\predCloNormImg_*$ is the front or back clothed normal map predicted by \camera{$\normG_\text{F,B}$ }
from $\{\inputimg, \bodyNormImg\}$, and $\bodyDepthImg_*$ is the front or back coarse body depth image rendered from the \smplx mesh, $\bodyMesh$.

Specifically, our objective function consists of five terms: 
\begin{align}
\begin{split}
    \min_{\clothDepthImgF, \clothDepthImgB} &\bniLoss(\clothDepthImgF;\predCloNormImgF) + 
    \bniLoss(\clothDepthImgB;\predCloNormImgB) + \\
    &\lambda_{\text{d}} \depthPriorLoss(\clothDepthImgF;\bodyDepthImgF) +
    \lambda_{\text{d}} \depthPriorLoss(\clothDepthImgB;\bodyDepthImgB) + \\
    &\lambda_{\text{s}} \silhouetteLoss(\clothDepthImgF, \clothDepthImgB)
    \text{,}
\end{split}
\label{eq.dbni_functional}
\end{align}
%\SUBSPRINT{where \bniLoss is front and back \bni terms introduced by \bni ~\cite{xu2022bilateral}, 
\camera{where \bniLoss is the \bni loss term introduced by \bni ~\cite{xu2022bilateral}, 
\depthPriorLoss is a depth prior applied to the front and back depth surfaces, and 
\silhouetteLoss is a front-back silhouette consistency term. }
\change{For a more detailed discussion on these terms, see \cref{sec: suppl-dbni} in~\suppl}

\begin{figure}[t]
    \centering
    % \vspace{-1.0 em}
    \includegraphics[width=1.0 \linewidth]{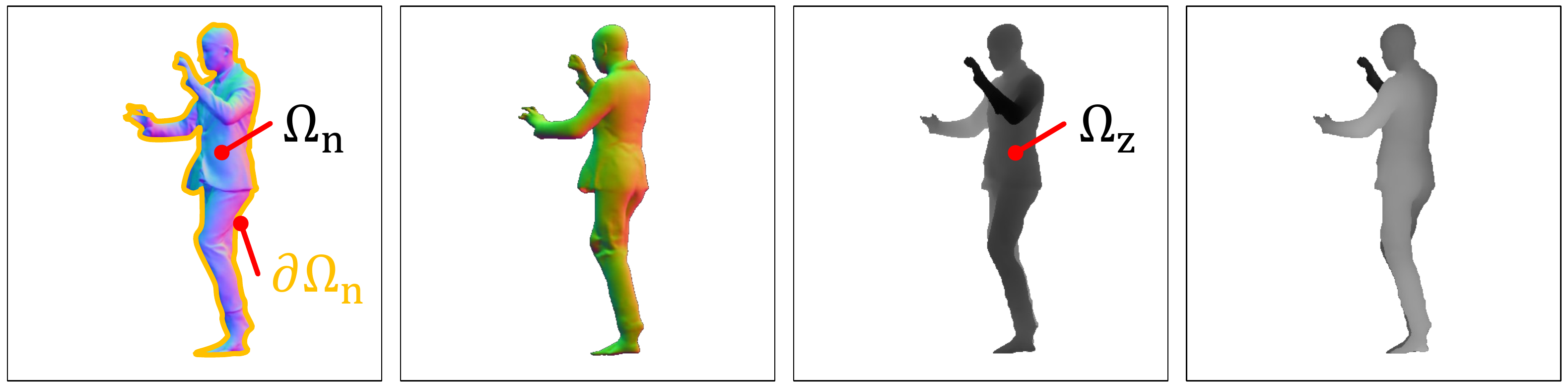}
    \begin{tabularx}{\linewidth}{
 >{\centering\arraybackslash}X
 >{\centering\arraybackslash}X
 >{\centering\arraybackslash}X
 >{\centering\arraybackslash}X
}
$\predCloNormImgF$ &
$\predCloNormImgB$ &
 $\bodyDepthImgF$ & 
 $\bodyDepthImgB$
 \end{tabularx}
    \caption{\camera{
    \cheading{Four inputs to \dbni} \domainN and \domainD are the domains of clothed and body regions, respectively. \silhouette is the silhouette of \domainN.}}
      \vspace{-1.5 em}
\label{fig:dbni_inputs}
\end{figure}

With \cref{eq.dbni_functional}, we make two technical contributions beyond \bni~\cite{xu2022bilateral}.
First, we use the coarse depth prior rendered from the \smplx body mesh, \SUBSPRINT{$\bodyDepthImg_i$}, to 
regularize
\bni:
\begin{align}
  \depthPriorLoss(\clothDepthImg_i;\bodyDepthImg_i) = | \clothDepthImg_i - \bodyDepthImg_i|_{\domainN \bigcap \domainD} \quad i \in \{F, B\} .
  \label{eq.dbni_depth}
\end{align}
This addresses the key problem of putting the front and back surfaces together in a coherent way to form a full body.
Optimizing \bni terms \bniLoss leaves an arbitrary global offset between the front and back surfaces.
The depth prior terms \depthPriorLoss encourage the surfaces with undecided offsets to be consistent with the \smplx body, and is computed in the domains 
$\domainN \bigcap \domainD$ (\cref{fig:dbni_inputs}). 
For further intuitions on \bniLoss and \depthPriorLoss, see~\cref{fig: dbni-K} and~\cref{fig: dbni-depth} in \suppl

Second, we use a silhouette consistency term to encourage the front and back depth values to be
the same at the silhouette boundary, which is computed in domain $\silhouette$ (\cref{fig:dbni_inputs}):
\begin{align}
    \silhouetteLoss(\clothDepthImgF, \clothDepthImgB) = 
    |\clothDepthImgF - \clothDepthImgB|_{\silhouette}.
    \label{eq.dbni_boundary}
\end{align}
The silhouette term improves the physical consistency of the reconstructed front and back clothed depth maps.
Without this term, \dbni~\change{produces} intersections of the front and back surfaces around the silhouette, \change{causing ``blobby'' artifacts \change{and hurting} reconstruction quality; see~\cref{fig: dbni-bc} in \suppl} 

\subsection{Human shape completion}
\label{sec: human shape completion}
For simple body poses without self-occlusions, merging front and back \dbni surfaces \change{in a straightforward way}, as done in \facsimile~\cite{smith2019facsimile} and \modulehuman~\cite{gabeur2019moulding}, can \change{result in} a \change{complete} \threeD clothed scan.
\change{However, often poses result in self-occlusions, which cause large portions of the surfaces to be missing.} 
In such cases, Poisson Surface Reconstruction (PSR)~\cite{kazhdan2006poisson} leads to blobby artifacts.

\begin{figure}
    \centering
    \includegraphics[trim=000mm 000mm 000mm 000mm, clip=true, width=1.0\linewidth]{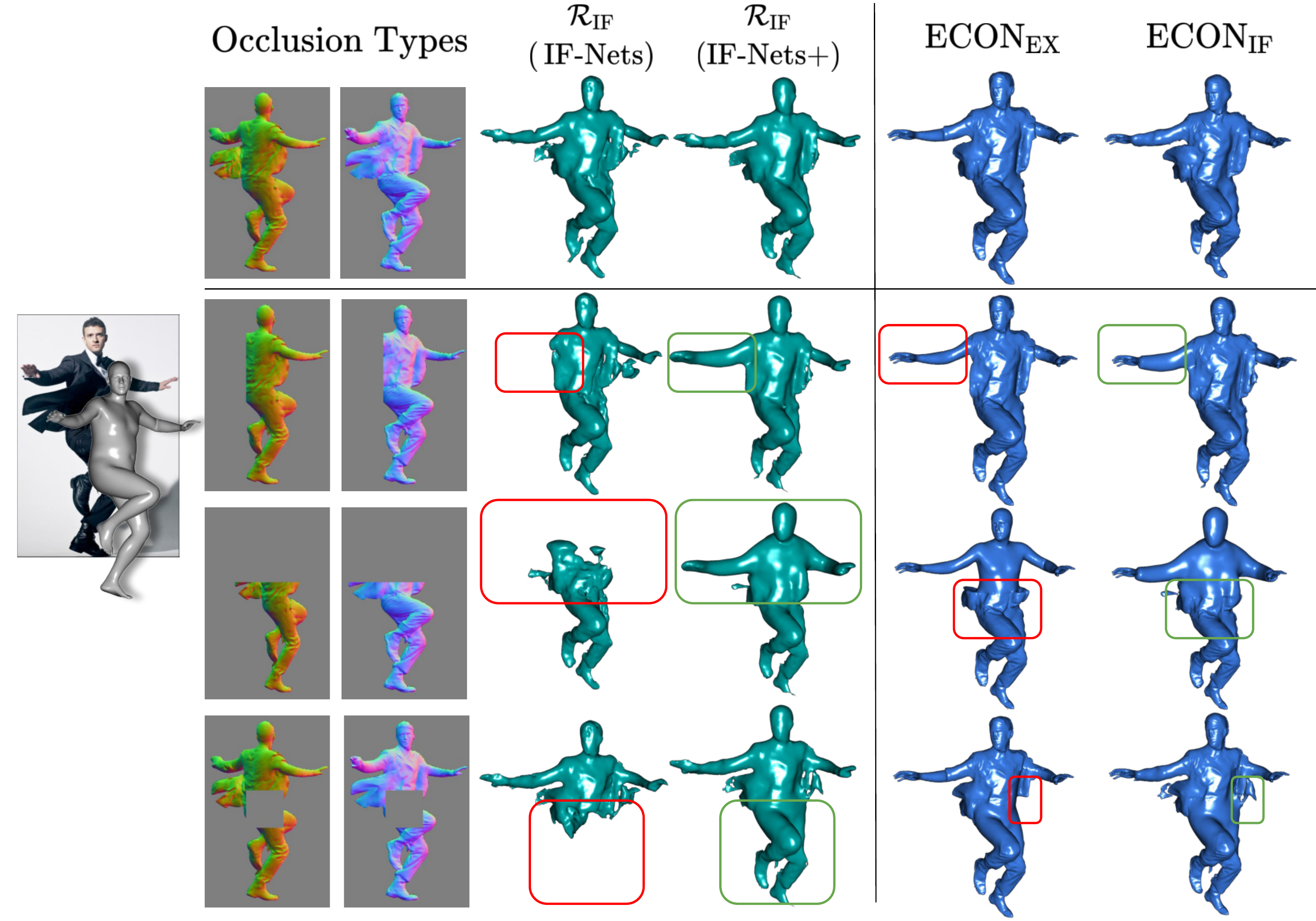}
    \caption{
                \cheading{``Inpainting'' the missing geometry}
                We simulate different cases of occlusion by masking the normal images and present the intermediate and final 3D reconstruction of different design choices. While \ifnet misses certain body parts, \ifnetplus produces a plausible overall shape. $\modelname_{\text{IF}}$ produces more consistent clothing surfaces than $\modelname_{\text{EX}}$ due to a learned shape distribution.
    }
    % \vspace{-1.5 em}
    \label{fig: complement}
\end{figure}

\myparagraph{\change{PSR} completion with \smplx ($\textcolor{frontcolor}{\modelname_\text{EX}}$).}
A naive way to ``infill'' the missing surface is to make use of the estimated \smplx body. 
We remove the triangles from $\bodyMesh$ that are visible to front or back cameras. The remaining triangle ``soup" $\cullbodyMesh$ contains both side-view boundaries and occluded regions. 
\change{We apply PSR~\cite{kazhdan2006poisson}} to the union of $\cullbodyMesh$ and \dbni surfaces $\{\FrontSurface, \BackSurface\}$ to obtain \change{a} watertight reconstruction $\recMesh$.
This approach is denoted as $\modelname_{\text{EX}}$. 
Although \change{$\modelname_{\text{EX}}$}
avoids missing limbs \change{or sides}, it does not produce a coherent surface for the originally missing  clothing and hair surfaces because of the discrepancy between \smplx and actual clothing or hair;
 see $\modelname_\text{EX}$ in~\cref{fig: complement}. 

\myparagraph{Inpainting with \ifnetplus ($\textcolor{GreenColor}{\recMeshif}$).} 
\change{To improve reconstruction coherence,
we use a learned implicit-function (\IF) model to ``inpaint'' the missing geometry given front and back \dbni surfaces.}
\change{Specifically, we tailor a general-purpose shape completion method, \ifnet~\cite{chibane20ifnet}, to a \smplx-guided one, denoted as \ifnetplus.}
\ifnet~\cite{chibane20ifnet} completes the \threeD shape from a deficient \threeD input, such as 
an incomplete %articulated 
3D human \change{shape} % (various representations)}
or a low-resolution voxel \change{grid}. 
\change{Inspired by Li~\etal~\cite{Li2022SHARP}}, we \change{adapt} \ifnet by conditioning it on a voxelized \smplx body to deal with pose variation; \change{for details see~\cref{sec: suppl-ifnetplus} in \suppl} 
\ifnetplus is trained 
\change{on voxelized front and back ground-truth clothed depth maps, $\{\gtclothDepthImgF, \gtclothDepthImgB\}$, and a voxelized (estimated) body mesh, $\bodyMesh$, as input,}
and is supervised with ground-truth 3D shapes. 
During training, we randomly mask \change{$\{\gtclothDepthImgF, \gtclothDepthImgB\}$} for robustness to occlusions.
\change{During inference, we feed the estimated 
% \dbni surfaces, $\clothDepthImgF$ and~$\clothDepthImgB$, and body mesh, $\bodyMesh$ 
$\clothDepthImgF$, $\clothDepthImgB$ and $\bodyMesh$ 
into \ifnetplus to obtain an occupancy field, from which we extract the \change{inpainted} mesh, $\recMeshif$, with \mcubes~\cite{lorensen1987marching}.}

\begin{figure}
    \centering
    \includegraphics[trim=000mm 000mm 000mm 000mm, clip=true, width=1.0\linewidth]{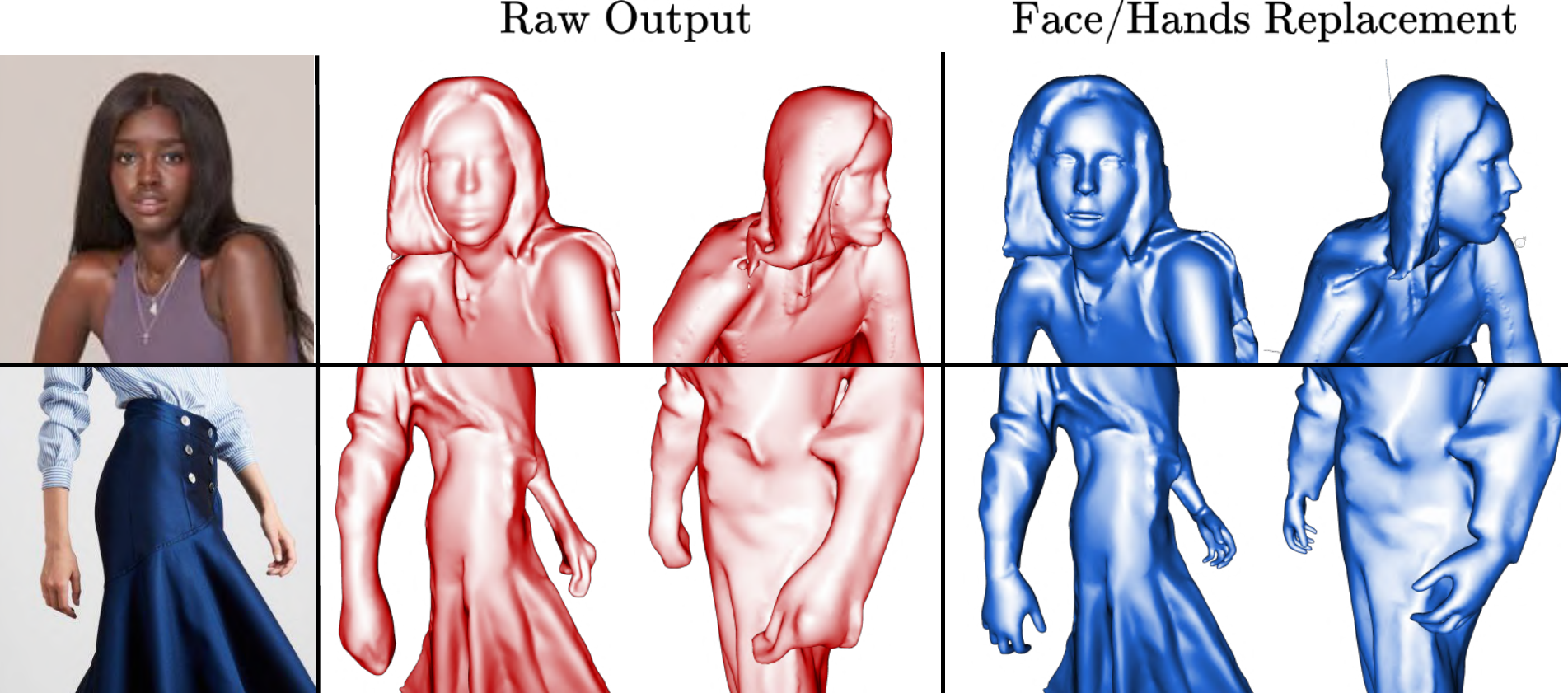}
    \caption{
                \cheading{Face and hand details}
                The face and hands of the raw reconstruction can be replaced with the ones of the \smplx body.
    }
    \label{fig: replacement}
    \vspace{-0.5 em}
\end{figure}

\myparagraph{\change{PSR} completion with $\recMeshif$ ($\textcolor{frontcolor}{\modelname_\text{IF}}$).}
To obtain our final mesh, $\recMesh$, we apply PSR to stitch (1) \dbni surfaces, (2) sided and occluded triangle soup $\cullbodyMesh$ from $\recMeshif$,  and optionally, (3) face or hands cropped from the estimated \smplx body $\bodyMesh$.
% This is due to the lossy voxelization of inputs and limited resolution of \mcubes; see the local difference between $\modelname_{\{\text{IF,EX}\}}$ and $\recMeshif$ in~\cref{fig: complement}.
% Further, we use the face or hands cropped from $\bodyMesh$ because these 
\camera{The necessity of (3) arises from the poorly reconstructed hands/face in $\recMeshif$, see difference in~\cref{fig: replacement}. The approach is denoted as $\modelname_{\text{IF}}$.}

% $\recMeshif$ significantly smooths out the details of $\BniDepths / \BniSurfaces$ optimized via \dbni (see $\textcolor{GreenColor}{\recMeshif}$ vs $\textcolor{frontcolor}{\modelname_\text{IF/EX}}$ in \cref{fig: complement}).
% $\text{\modelname}_{\text{IF/EX}}$ preserves \dbni details better, since only the side-views and occluded parts of $\recMeshif$ are fused in the Poisson step.

\camera{Notably, although $\recMeshif$ is already a complete human mesh, 
% we only use its side and occluded parts because its front and back regions lack \change{sharp} details compared to \dbni surfaces.
due to the lossy voxelization of inputs and limited resolution of \mcubes algorithm, it somehow smooths out the details of $\BniDepths / \BniSurfaces$, which are optimized via \dbni (see $\textcolor{GreenColor}{\recMeshif}$ vs $\textcolor{frontcolor}{\modelname_{\{\text{IF,EX}\}}}$ in \cref{fig: complement}). While $\modelname_{\{\text{IF,EX}\}}$ preserves \dbni details better, only the side-views and occluded parts of $\recMeshif$ are fused in the Poisson step. In \cref{tab:benchmark,tab:benchmarkShapeCOMPL}, we use $\modelname_{\{\text{IF,EX}\}}$ instead of $\recMeshif$ for evaluation}.

\section{Experiments}
\label{sec: experiments}

\subsection{Datasets} % and metrics}
% \label{sec: dataset and metrics}
\label{sec: dataset}

\qheading{Training on \thuman} %as train set}
\thuman~\cite{tao2021function4d} contains 525 high-quality human textured scans in various poses, which are captured by a dense DSLR rig, along with their corresponding \smplx fits. We use \thuman to train \icon, $\text{ECON}_\text{IF}$ (\ifnetplus), \ifnet, \pifu and \pamir.

\smallskip
\qheading{Quantitative evaluation on \cape \& \renderppl} %as test set}
We primarily evaluate on \cape \cite{ma2020cape} and \renderppl \cite{renderpeople}. Specifically, we use the \capeNFP set (100 scans), which is used by \icon to analyze  robustness to complex human poses. Moreover, we select another 100 scans from \renderppl, containing loose clothing, such as dresses, skirts, robes, down jackets, costumes, \etc. 
\SUBSPRINT{With such clothing variance, \renderppl helps numerically evaluate the flexibility of reconstruction methods \wrt shape topology.}
Samples of the two datasets are shown in \cref{fig: dataset}.

\begin{figure}[t] %[h]
    \centering
    \includegraphics[trim=000mm 000mm 000mm 000mm, clip=true, width=1.00\linewidth]{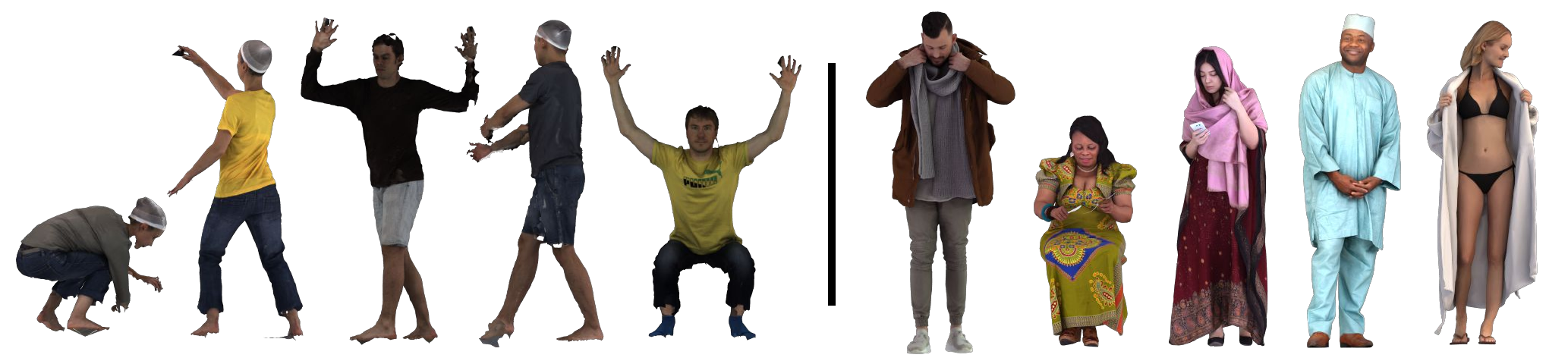}
    \caption{
                \cheading{Datasets for numerical evaluation}
                We evaluate \modelname on images with unseen poses (left) and unseen outfits (right) on the \cape~\cite{ma2020cape} and \renderppl~\cite{renderpeople} datasets, respectively.
    }
    % \vspace{-1.0em}
     \label{fig: dataset}
\end{figure}

\subsection{Metrics}
\label{sec: metrics}

\qheading{Chamfer and P2S distance (cm)}
\camera{To capture large geometric errors, \eg occluded parts or wrongly positioned limbs}, we report the \SUBSPRINT{commonly used} 
Chamfer (bi-directional point-to-surface) and 
P2S distance (1-directional point-to-surface) between \groundtruth %scans 
and reconstructed meshes. 

\qheading{Normal difference (L2)}
\camera{To measure the fineness of reconstructed local details, as well as projection consistency from the input image}, we also report the L2 error between normal images rendered from reconstructed and \groundtruth surfaces, by rotating a virtual camera around these by
$\{0^\circ, 90^\circ, 180^\circ, 270^\circ\}$ \wrt to a frontal view. 
\smallskip

\subsection{Evaluation}
\label{sec: evaluation}

\begin{table}[t]
\centering{
\resizebox{\linewidth}{!}{  %   \cite{ma2020cape}  \cite{renderpeople}
  \begin{tabular}{c|c|ccc|ccc}
    Methods & Data-driven & \multicolumn{3}{c|}{\OOD poses (\cape)} 
            & \multicolumn{3}{c}{\OOD outfits (\renderppl)}\\
     &  &Chamfer $\downarrow$ & P2S $\downarrow$ & Normals $\downarrow$ 
     & Chamfer $\downarrow$ & P2S $\downarrow$ & Normals $\downarrow$\\
    \shline
    \multicolumn{7}{c}{w/o \smplx body prior} \\
    \hline
    $\text{\pifu*}$ & \cmark & 1.722 & 1.548 & 0.0674 & 1.706 & 1.642 & 0.0709\\
    $\text{\pifuhd}^\dagger$ & \cmark & 3.767 & 3.591 & 0.0994 & 1.946 & 1.983 & 0.0658 \\
    \hline
    \multicolumn{7}{c}{w/ GT \smplx body prior} \\
    \hline
    $\text{\pamir*}$ & \cmark & 0.989 & 0.992 & 0.0422 & \bone 1.296 & \btwo 1.430 &  0.0518 \\
    % $\text{\icon-w/o F}$~\cite{xiu2022icon} & T  & 1.040 & 1.016 & 0.0427 & 1.763 & 2.192 & 0.0732 \\
    $\text{\icon}$ & \cmark & \btwo 0.971 & \bone 0.909 & \btwo 0.0409 & 1.373 & 1.522 & 0.0566 \\
    \hline
    $\text{\modelname}_{\text{IF}}$ & \cmark &  0.996 & 0.967 & 0.0413 & 1.401 & \bone 1.422 & \btwo 0.0516 \\
    $\text{\modelname}_{\text{EX}}$ & \xmark & \bone 0.926 & \btwo 0.917 & \bone 0.0367 & \btwo 1.342 & 1.458 & \bone 0.0478 \\
  \end{tabular}}
}
\caption{
    \cheading{Evaluation against the state of the art}
    All models use a resolution of 256 for marching cubes. 
    $^*$Methods are re-implemented in \cite{xiu2022icon} 
    \SUBSPRINT{for a fair comparison in terms of network settings and training data.} 
    $^\dagger$Official model is trained on the \renderppl dataset. \camera{$\text{\modelname}_{\text{EX}}$ is optimization-based, thus requires no training (\xmark). ``OOD'' is short for ``out-of-distribution''.}
}
\label{tab:benchmark}
\end{table}

\qheading{Quantitative evaluation}
We compare \modelname
with body-agnostic methods, \ie, \pifu~\cite{saito2019pifu} and~\pifuhd~\cite{saito2020pifuhd}, and body-aware methods, \ie, \pamir~\cite{zheng2020pamir} and~\icon~\cite{xiu2022icon}; see in~\cref{tab:benchmark}. 
For fair comparison, we use re-implementations of \pifu and \pamir from \icon~\cite{xiu2022icon}, 
because they have the same network settings and input data.
\camera{$\modelname_\text{EX}$ performs on par with \icon, and outperforms other methods on images containing \ood (OOD) poses (\cape), with a distance error below 1cm.} 
In terms of \ood outfits (\renderppl), $\modelname_\textbf{EX/IF}$ performs on par with \pamir, and much better than \pifuhd. 
\camera{When it comes to high-frequency details measured by normals, 
% both variants of \modelname ($\modelname_\text{IF}$, $\modelname_\text{EX}$), 
$\modelname_\text{EX}$ achieves \sota performance on both datasets.}

% \begin{table}%[H]
% \centering
% % \scriptsize
% \small
% % \resizebox{\linewidth}{!}
% {
% \begin{tabular}{c|c|c}
%  & \icon~\cite{xiu2022icon} & \pifuhd~\cite{saito2020pifuhd}\\
% \shline
% % \multirow{2}{*}{Difficult poses} & Preference \\ P-value & 0.423 \\ 0.182
% % \hline
% % \multirow{2}{*}{Loose clothing} & Preference \\ 0.213 & 0.457
% % \hline
% % \multirow{2}{*}{Fashion images} & Preference \\ 0.335 & 0.519
% Challenging poses &  \textcolor{AlphaColor}{0.423} & \textcolor{AlphaColor}{0.182} \\
% Loose clothing    &  \textcolor{AlphaColor}{0.213} & \textcolor{AlphaColor}{0.457} \\
% Fashion images    &  \textcolor{AlphaColor}{0.335} & \textcolor{RedColor}{0.519} \\
% \end{tabular}
% }
% \caption{
%             \cheading{Perceptual study} 
%             Numbers denote the chance that participants prefer the reconstruction of a competing method over \modelname for \inthewild images. 
%             A value of $0.5$ indicates equal preference.
%             A value of \textcolor{AlphaColor}{$<0.5$  favors ECON}, while 
%                     of \textcolor{RedColor}{$>0.5$    favors competitors}. 
%             % \modelname is judged significantly more realistic. 
% }
% \label{tab: perceptual study}
% \end{table}

\begin{table}%[H]
\centering
% \scriptsize
\small
% \resizebox{\linewidth}{!}
{
\begin{tabular}{c|c|c|c}
 & \icon~\cite{xiu2022icon} & \pifuhd~\cite{saito2020pifuhd} &
 \pamir~\cite{zheng2020pamir}\\
\shline
% \multirow{2}{*}{Difficult poses} & Preference \\ P-value & 0.423 \\ 0.182
% \hline
% \multirow{2}{*}{Loose clothing} & Preference \\ 0.213 & 0.457
% \hline
% \multirow{2}{*}{Fashion images} & Preference \\ 0.335 & 0.519
Challenging poses &  \textcolor{AlphaColor}{0.283} & \textcolor{AlphaColor}{0.108} & \textcolor{AlphaColor}{0.132}\\
Loose clothing    &  \textcolor{AlphaColor}{0.147} & \textcolor{AlphaColor}{0.362} & \textcolor{AlphaColor}{0.232}\\
Fashion images    &  \textcolor{AlphaColor}{0.199} & \textcolor{RedColor}{0.551} & \textcolor{AlphaColor}{0.290}\\
\end{tabular}
}
\caption{
            \cheading{Perceptual study} 
            Numbers denote the chance that participants prefer the reconstruction of a competing method over \modelname for \inthewild images. 
            A value of $0.5$ indicates equal preference.
            A value of \textcolor{AlphaColor}{$<0.5$  favors ECON}, while 
                    of \textcolor{RedColor}{$>0.5$    favors competitors}. 
            % \modelname is judged significantly more realistic. 
}
\label{tab: perceptual study}
\end{table}

\smallskip
\qheading{Perceptual study} Due to the lack of ground-truth geometry (clothed scan + underneath \smplx), we further conduct a perceptual study to evaluate \modelname on \itw images. 
Test images are divided into three categories: 
``challenging poses'', 
``loose clothing'', and 
``fashion images''.
Examples of challenging poses and loose clothing can be seen in \cref{fig: qualitative figure}, and some of fashion images are in~\suppl's \cref{fig: SHHQ}.

% Due to the lack of ground-truth geometry for in-the-wild images (clothed scan + underneath \smplx), we perform a perceptual study. % on the results. 
Participants are asked to choose the reconstruction they perceive as more realistic, 
between 
a baseline method and \modelname. 
We compute the chances that each baseline is preferred over \modelname in~\cref{tab: perceptual study}. 
The results of the perceptual study confirm
the  quantitative evaluation in~\cref{tab:benchmark}.
For ``challenging poses'' images, \modelname is significantly preferred over \mbox{PIFuHD} and outperforms \mbox{ICON}. 
On images of people wearing loose clothing, \modelname is preferred over ICON by a large margin and outperforms \mbox{PIFuHD}. 
The reasons for a slight preference of \pifuhd over \modelname on fashion images are discussed in~\cref{sec:limitation}.
%This confirms that, \modelname combines \dbni 
%with a body prior in a highly effective way; 
%on the one hand, it is robust to unseen poses, while 
%on the other hand, it is capable of reconstructing loose clothing and geometric details, 
%as reconstructed shape is not 
%over-constrained to the topology of a \smplx body. %body topology but 
\Cref{fig:motivation} visualizes some comparisons. 
More examples are provide in \cref{fig: result-pose,fig: result-cloth,fig: result-fashion} of the \suppl

\begin{table}[t] %[H]
\centering{
\resizebox{0.48\textwidth}{!}{  %   \cite{ma2020cape}  \cite{renderpeople}
  \begin{tabular}{@{}l|cc|cc|c}
    Methods & \multicolumn{2}{c|}{\OOD poses (\cape \cite{ma2020cape})} & \multicolumn{2}{c|}{\OOD outfits (\renderppl) \cite{renderpeople}} & Speed \\
      & RMSE $\downarrow$ & MAE $\downarrow$ & RMSE $\downarrow$ & MAE $\downarrow$ & FPS $\uparrow$\\
    \shline
    $\text{\bni}$~\cite{xu2022bilateral} & 27.64 & 21.11 & 20.61 & 16.07 & 0.52 \\
    $\text{\dbni}$ &  \textbf{13.43} &  \textbf{10.29} &  \textbf{14.43} &  \textbf{11.26} &  \textbf{0.69}\\
  \end{tabular}}
}
\vspace{-0.7 em}
\caption{
    \cheading{\bni vs \dbni}
    Comparison between \bni and \dbni surfaces \wrt reconstruction accuracy and optimization speed.
}
\label{tab:benchmarkBNI}
\end{table}

\subsection{Ablation study}
\qheading{\dbni vs \bni}
\label{sec: compare normal integration}
We compare \dbni with \bni using $600$ samples (200 scans x 3 views) from \cape and \renderppl where ground-truth normal maps and meshes are available.
\Cref{tab:benchmarkBNI} reports the 
``root mean squared error'' (RMSE) and 
``mean absolute error'' (MAE) 
between the estimated and rendered depth maps.
\dbni significantly improves the reconstruction accuracy by about $50\%$ compared to \bni.
This demonstrates the efficacy of 
using the coarse body mesh as regularization and 
taking the consistency of both the front and back surface into consideration.
Additionally, \dbni is $33 \%$ faster than \bni.

\qheading{\ifnetplus vs \ifnet}
\label{sec: compare ifnet and ifnetplus}
Following the \change{metrics of} \cref{sec: metrics}, we compare \ifnet~\cite{chibane20ifnet} with our \ifnetplus on $\recMeshif$. 
We show the quantitative comparison in \cref{tab:benchmarkShapeCOMPL}. 
The
improvement for 
\ood (``OOD'') poses shows that \ifnetplus is more robust to pose variations than \ifnet, as it is conditioned on the \smplx body.
\Cref{fig: complement} \change{compares the} %illustrates the comparison of 
geometry ``inpainting'' of both methods in the case of occlusions. 

\begin{table}[t] %[H]
\centering{
\resizebox{0.48\textwidth}{!}{
  \begin{tabular}{c|ccc|ccc}
    Methods & \multicolumn{3}{c|}{OOD poses (\cape)} 
            & \multicolumn{3}{c}{OOD outfits (\renderppl)}\\
     & Chamfer $\downarrow$ & P2S $\downarrow$ & Normals $\downarrow$ 
     & Chamfer $\downarrow$ & P2S $\downarrow$ & Normals $\downarrow$\\
    \shline
    \ifnet~\cite{chibane20ifnet} & 2.116 & \textbf{1.233} & 0.075 & 1.883 & 1.622 & 0.070 \\
    \ifnetplus & \textbf{1.401} & 1.353 & \textbf{0.056} & \textbf{1.477} & \textbf{1.564} & \textbf{0.055} \\
    \hline
    $\text{\modelname}_{\text{IF}}$ & 0.996 & 0.967 & 0.0413 & 1.401 & 1.422 & 0.0516 \\
  \end{tabular}}
}
\vspace{-0.7 em}
\caption{
    \cheading{Evaluation for shape completion} Same metrics as~\cref{tab:benchmark}, and $\modelname_\text{IF}$ is added as a reference.
}
% \vspace{-0.5 em}
\label{tab:benchmarkShapeCOMPL}
\end{table}

\begin{figure}[tb] %[H]
    \includegraphics[trim=000mm 000mm 000mm 000mm, clip=true, width=1.0\linewidth]{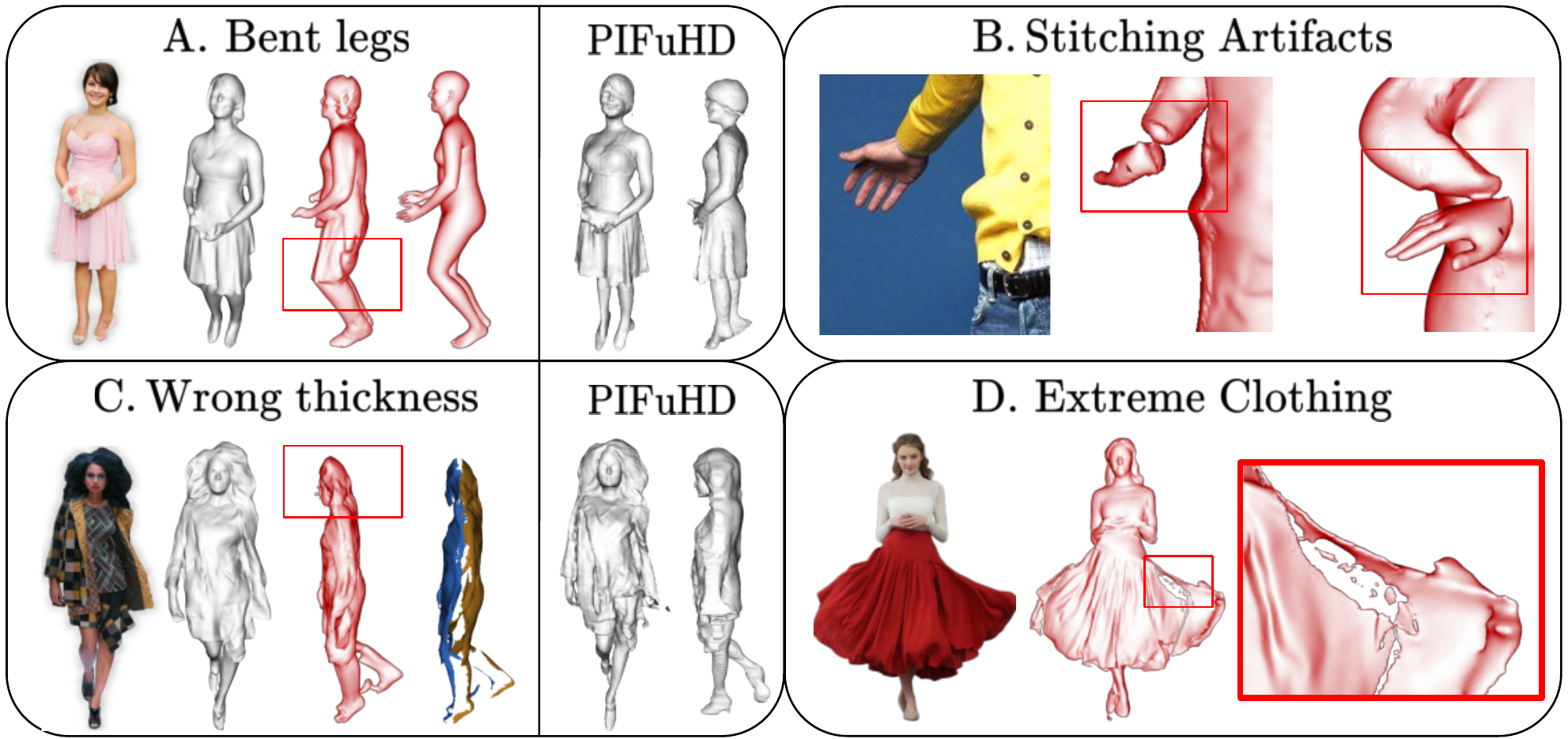}
    \vspace{-0.7 em}
    \caption{
                \cheading{Failure examples of \modelname}
                    (A-B)   Failures in recovering a \smplX body result, \eg, 
                            bent legs or wrong limb poses, 
                            cause \modelname failures by extension. 
                    (C-D)   Failures in normal-map estimation 
                            provide erroneous geometry to \modelname 
                            to work with. 
    }
    % \vspace{-0.8 em}
    \label{fig: failure cases}
\end{figure}

\begin{figure*}[t]
\vspace{-2em}
\centering{
    \begin{subfigure}{\linewidth}
        \includegraphics[trim=000mm 000mm 000mm 000mm, clip=true, width=1.0\linewidth]{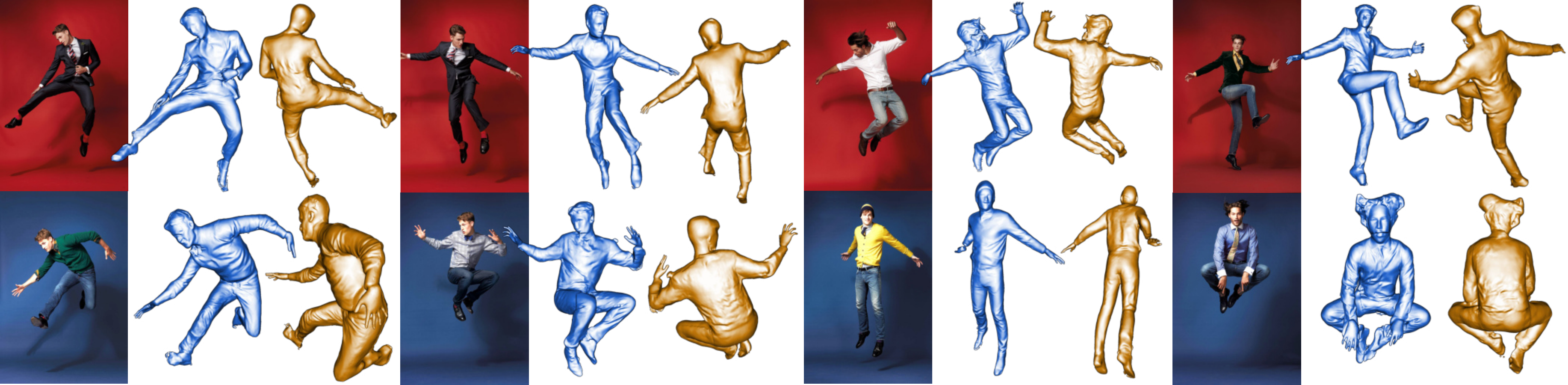}
        \caption{
        Humans with challenging poses (\faSearch~\textbf{zoom in} to see \threeD wrinkle details)
        }
        \label{fig: challenging pose}
    \end{subfigure}
    \begin{subfigure}{\linewidth}
        \includegraphics[width=\linewidth]{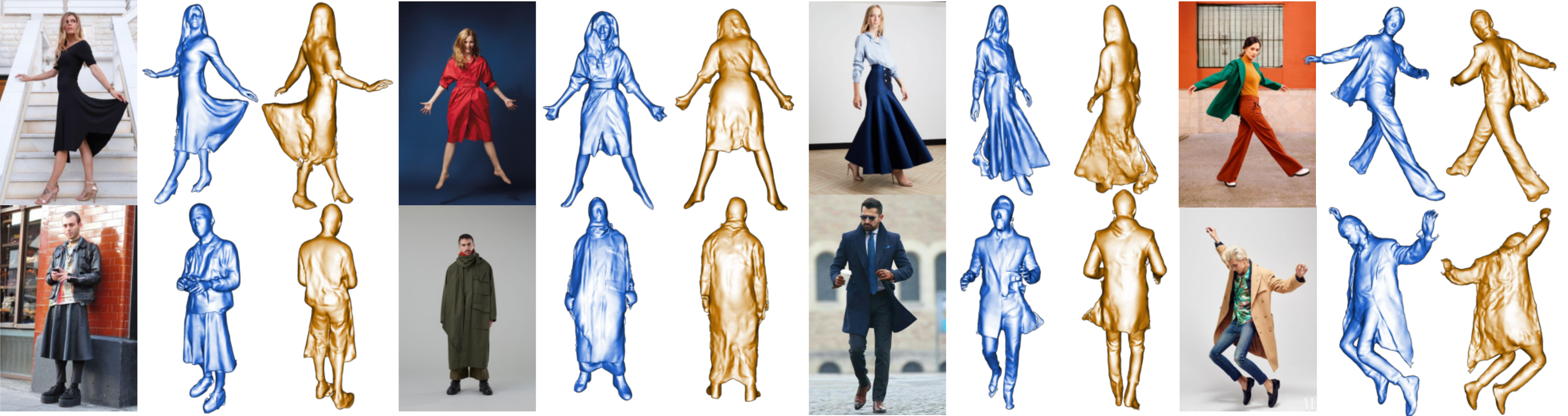}
	    \caption{
	    Humans with loose clothing (\faSearch~\textbf{zoom in} to see \threeD clothing details)
	    }
	    \label{fig: loose clothing}
    \end{subfigure}}
    \vspace{-1.0em}
    \caption{
                \cheading{Qualitative results on \itw images}
                We show 8 examples of reconstructing 
                detailed clothed \threeD humans
                from images with: 
                (a) challenging poses and 
                (b) loose clothing. % respectively. 
                For each example we show the input image along with two
                views (\textbf{\textcolor{frontcolor}{front}} and 
                \textbf{\textcolor{backcolor}{rotated}})
                of the reconstructed \threeD humans. 
                Our approach is robust to pose variations, generalizes well to loose clothing, and contains detailed geometry. 
    }
    \vspace{-2.0mm}
\label{fig: qualitative figure}
\end{figure*}

%\section{Applications}
%\label{sec: applications}

%\qheading{Multi-person reconstruction}
\subsection{Multi-person reconstruction}
\label{sec: occluded-person}
Thanks to the shape completion module, \modelname can deal with occlusions. 
Unlike other crowd body estimators~\cite{sun2021BEV,sun2023trace,sun2021ROMP,ye2023slahmr}, \modelname makes it possible to reconstruct multiple detailed ``clothed'' \threeD humans from an image with inter-person occlusions, 
even though \modelname has not been trained for this.
\Cref{fig: crowd} shows three examples. 
The occluded parts, colored in red, are successfully recovered. 

\begin{figure*}[t]
    \includegraphics[trim=000mm 000mm 000mm 000mm, clip=true, width=1.0\linewidth]{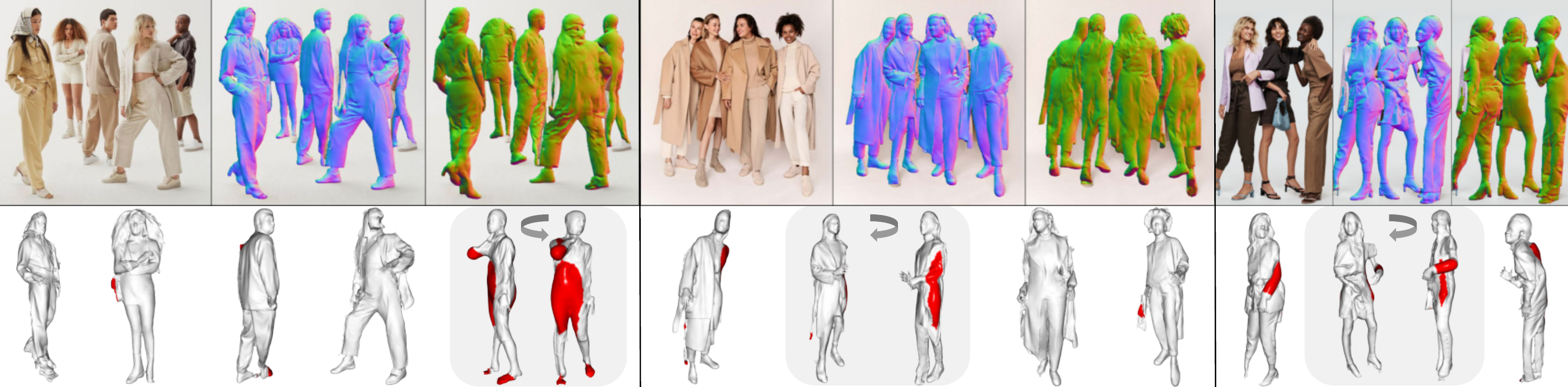}
    \caption{
                \cheading{Multiple humans with occlusions.}
                We detect multiple people and apply \modelname to each separately. 
                Although \modelname is not trained on multiple people, it is robust to inter-person occlusions.
                We show three examples, and for each: 
                (top)
                input image and the predicted front and back normal maps, 
                (bottom)
                \modelname's reconstruction.
                Red areas on the estimated mesh indicate occlusions. 
    }
    \vspace{-1.0em}
    \label{fig: crowd}
\end{figure*}

\iffalse
\smallskip
\qheading{Augment \twoD human datasets with \modelname}
\label{sec: multi-person}
Datasets of real clothed humans with \threeD ground truth \cite{renderpeople, ma2020cape, axyz, tao2021function4d, humanalloy, zheng2019deephuman} are limited in size. 
%
In contrast, datasets of images without \threeD ground truth are widely available in large sizes~\cite{fu2022stylegan, DeepFashion2,liu2016deepfashion}. 
%
We can ``augment'' such datasets by reconstructing detailed 3D humans from their images. 
We apply \modelname on SHHQ~\cite{fu2022stylegan} and recover normal maps and 3D humans. 
\Cref{fig: SHHQ} shows some examples. 
%
\camera{As \modelname-like methods mature, they could produce pixel-aligned \threeD humans from photos on large-scale}, to help people train pose regressors, or generate clothed avatars with details \cite{zhang2022avatargen,EVA3D,Chan2022eg3d,grigorev2021stylepeople,noguchi2022unsupervised,bergman2022gnarf,jiang2022humangen}. 
\fi

\section{Discussion}
\label{sec:limitation}

\qheading{Limitations}
\modelname takes as input an \rgb image and an estimated \smplX body. 
However, 
recovering \smplX bodies (or similar models) 
from a single image is still an open problem, and not fully solved. 
Any failure in this could lead to \modelname failures, such as in 
\mbox{\cref{fig: failure cases}\textcolor{red}{-A}} and 
\mbox{\cref{fig: failure cases}\textcolor{red}{-B}}. \camera{As the synthetic data~\cite{black2023bedlam,hewitt2023procedural,weitz2021infiniteform} is getting sufficiently realistic, their domain gap with real data is significantly narrowed, it is predictable that such limitations will be eliminated.} 
The reconstruction quality of \modelname primarily relies on the accuracy of the predicted normal maps. 
Poor normal maps can result in overly close-by or even intersecting front and back surfaces, as shown in 
\mbox{\cref{fig: failure cases}\textcolor{red}{-C}} and 
\mbox{\cref{fig: failure cases}\textcolor{red}{-D}}. 

\smallskip
\qheading{Future work}
Apart from addressing the above limitations, several other directions are useful for practical applications. 
Currently, \modelname reconstructs only 3D geometry. 
One could additionally recover an underlying skeleton and skinning weights~\cite{RigNet,li2021learning}, to obtain fully-animatable avatars. 
Moreover, generating back-view texture~\cite{zhang2023adding,rombach2022high,richardson2023texture,chen2023text2tex} would result in fully-textured avatars. 
Disentangling clothing~\cite{zhu2022registering,aggarwal2022layered,qiu2023REC-MV}, hairstyle~\cite{zheng2023hair}, or accessories~\cite{gao2022dart} from the recovered geometry, would enable the simulation~\cite{grigorev2023hood}, synthesis, editing and transfer of styles~\cite{feng2022scarf} for these. 
\modelname's reconstructions, together with its underneath \smplx body, could be useful as \threeD shape prior to learn neural avatars~\cite{dongtotalselfscan,jiang2022selfrecon,guo2023vid2avatar}.

%\smallskip 
%\qheading{Augment \twoD human datasets with \modelname}
%\label{sec: multi-person}
In particular, \modelname could be used to augment existing datasets of \twoD images with \threeD humans.
Datasets of real clothed humans with \threeD ground truth \cite{renderpeople,ma2020cape, patel2021agora,tao2021function4d,zheng2019deephuman} are limited in size. 
In contrast, datasets of images without \threeD ground truth are widely available in large sizes~\cite{fu2022stylegan, DeepFashion2,liu2016deepfashion}. 
We can ``augment'' such datasets by reconstructing detailed 3D humans from their images. 
In \suppl, we apply \modelname on SHHQ~\cite{fu2022stylegan} and recover normal maps and 3D humans; see \cref{fig: SHHQ}.
\camera{As \modelname-like methods mature, they could produce pixel-aligned \threeD humans from photos at scale}, enabling the training of generative models of \threeD clothed avatars with details \cite{zhang2022avatargen,EVA3D,gao2022get3d,grigorev2021stylepeople,noguchi2022unsupervised,jiang2022humangen,xiong2023get3dhuman}. 

\smallskip
\qheading{Possible negative impact}
As the reconstruction matures, it opens the potential for low-cost realistic avatar creation. Although such a technique  benefits entertainment, film production, tele-presence and future metaverse applications, it could also facilitate deep-fake avatars.
Regulations must be established to clarify the appropriate use of such technology.

\section{Conclusion}
\label{sec:conclusion}

We propose \modelname to reconstruct detailed clothed 3D humans from a color image.
\camera{It combines estimated \twoFiveD front and back surfaces 
with and underlying \threeD parametric body in a highly effective way.
On the one hand, it is robust to novel poses, while 
on the other hand, it is capable of recovering loose clothing and geometric details, 
since the reconstructed shape is not 
over-constrained to the topology of the body.}
% \modelname combines the best properties of explicit regularization and free-from representation; 
% it estimates detailed 3D surfaces for the human body and clothing 
% \camera{without being limited to specific topology, while being robust to challenging poses and clothing. }
\modelname achieves this by using and extending recent advances in variational normal integration~\cite{xu2022bilateral} and shape completion~\cite{chibane20ifnet}.
It effectively extends these to the task of image-based \threeD human reconstruction.
We believe \modelname can lead to both real-world applications and useful tools for the \threeD vision community. The code and models are available at \projectURL~for research purposes.

% \clearpage
%\pagebreak

% \smallskip
% \noindent\rule[0.5ex]{\linewidth}{1.5pt}

\qheading{Acknowledgments} 
We thank Lea Hering and Radek Daněček for proofreading, Yao Feng, Haven Feng and Weiyang Liu for valuable feedback, and Tsvetelina Alexiadis for perceptual study. \camera{We are especially grateful to Carlos Barreto (\href{https://carlosedubarreto.gumroad.com/l/CEB_ECON}{Blender Add-on}), Teddy Huang (\href{https://github.com/YuliangXiu/ECON/blob/master/docs/installation-docker.md}{Docker Image}), and Justin John (\href{https://github.com/YuliangXiu/ECON/blob/master/docs/installation-windows.md}{Windows Support}).}  
This project has received funding from the European Union’s Horizon $2020$ research and innovation programme under the Marie Skłodowska-Curie grant agreement No.$860768$ (\href{https://www.clipe-itn.eu}{CLIPE} project).

\qheading{Disclosure}
\href{https://files.is.tue.mpg.de/black/CoI_CVPR_2023.txt}{is.tue.mpg.de/black/CoI\_CVPR\_2023.txt}
% MJB has received research gift funds from Adobe, Intel, Nvidia, Meta/Facebook, and Amazon.  MJB has financial interests in Amazon, Datagen Technologies, and Meshcapade GmbH.  While MJB is a consultant for Meshcapade, his research in this project was performed solely at, and funded solely by, the Max Planck Society.

% \smallskip
% \noindent\rule[0.5ex]{\linewidth}{1.5pt}

% \renewcommand{\thefigure}{A.\arabic{figure}}
% \renewcommand{\thetable}{A.\arabic{table}}
% \renewcommand{\theequation}{A.\arabic{equation}}

\renewcommand{\thefigure}{S.\arabic{figure}}
\renewcommand{\thetable}{S.\arabic{table}}
\renewcommand{\theequation}{S.\arabic{equation}}

\setcounter{figure}{0}
\setcounter{table}{0}
\setcounter{equation}{0}

\begin{appendices}
\label{appendices}

In the following, we provide more details and  discussion on normal prediction, \dbni and \ifnetplus, as well as more qualitative results in the perceptual study, as an extension of \cref{sec: method} and \cref{sec: experiments} 
%\cref{sec: applications} 
of the main paper.
We also explore future applications.
Please check the \video for an overview of the method and more results.

\section{Implementation details}
\subsection{Normal map prediction}
\label{sec: suppl-normal-prediction}
We set the loss weights 
$\lambda_\text{\joint\text{\_diff}}$, 
$\lambda_\text{\normal\text{\_diff}}$, and 
$\lambda_\text{\mask\text{\_diff}}$ 
in~\cref{eq:body-fit} to 5.0, 1.0, and 1.0 respectively. However, if the overlap ratio between clothing and body mask is smaller than 0.5, it means humans are dressed with loose clothing. In this situation we trust the 2D joints more and increase the $\lambda_\text{\joint\text{\_diff}} = 50.0$. Similarly, when the overlap between body mask inside the clothing mask and full body mask is smaller than 0.98, occlusion happens. In such cases we set $\lambda_\text{\mask\text{\_diff}} = 0.0$ to avoid limb self-intersection after pose refinement.

During inference, following \icon~\cite{xiu2022icon}, we iteratively refine \smplx and clothed-body normals for 50 iterations (1.10 iter/s on Quadro RTX 5000 GPU). We use \specific{rembg}\footnote{\url{https://github.com/danielgatis/rembg}} plus \specific{\maskrcnn (ResNet50-FPN-V2)}~\cite{he2017mask} for multi-person segmentation, \specific{Mediapipe}~\cite{lugaresi2019mediapipe} to estimate full-body landmarks, \camera{\specific{Open3D}} for poisson surface reconstruction~\cite{kazhdan2006poisson}, and \specific{MonoPort}~\cite{li2020monoportRTL,li2020realtimeVolMoCap} for fast implicit surface query, and \specific{PyTorch3D}~\cite{ravi2020pytorch3d} for marching cubes.

\subsection{\dbni}
\label{sec: suppl-dbni}
\qheading{Optimization details}
To better present the optimization details, we first write the \dbni objective function in a matrix form.
\Cref{fig:dbni_inputs} shows the four inputs to \dbni.
We vectorize the front and back clothed and prior depth maps $\{\clothDepthImgF, \clothDepthImgB, \bodyDepthImgF, \bodyDepthImgB\}$ within \domainN as $\{\depthFront, \depthBack, \depthFrontPrior, \depthBackPrior\}$; all vectors are of length $|\domainN|$.
\dbni then jointly solves for the front and back clothed depth \depthFront and \depthBack by minimizing the objective function consisting of the five terms:
\begin{equation}
\begin{aligned}
    \mathcal{L}(\depthFront, \depthBack) =(\V{A}_\text{F}\depthFront - \V{b}_\text{F})^\top&\V{W}_{\text{F}}(\V{A}_\text{F}\depthFront - \V{b}_\text{F}) \\
    + (\V{A}_\text{B}\depthBack - \V{b}_\text{B})^\top&\V{W}_{\text{B}}(\V{A}_\text{B}\depthBack - \V{b}_\text{B}) \\
    + \lambda_\text{d}(\depthFront - \depthFrontPrior)^\top &\V{M} (\depthFront - \depthFrontPrior) \\
    + \lambda_\text{d}(\depthBack - \depthBackPrior)^\top &\V{M} (\depthBack - \depthBackPrior) \\
    + \lambda_\text{s}(\depthFront - \depthBack)^\top &\V{S} (\depthFront - \depthBack).
\end{aligned}
\label{eq: dbni_opt_long}
\end{equation}
Here, $\V{A}_\text{F} \in \mathbb{R}^{4|\domainN| \times |\domainN|}$ and  $\V{b}_\text{F} \in\mathbb{R}^{4|\domainN|}$ are constructed from the front normal map following \mbox{Eq. (21)} of \bni~\cite{xu2022bilateral}; $\V{A}_\text{B}$ and $\V{b}_\text{B}$ are from the back normal map.
$\V{W}_{\text{F}}$ and $\V{W}_{\text{B}} \in \mathbb{R}^{4|\domainN| \times 4|\domainN|}$ are bilateral weight matrices for front and back depth maps, respectively; both are constructed following \mbox{Eq.~(22)} of \bni~\cite{xu2022bilateral} and depend on the unknown depth.
$\V{M}$ and $\V{S}$ are $|\domainN| \times |\domainN|$ diagonal matrices whose diagonal entries indicate the pixels with depth priors and located at the silhouette, respectively.
Specifically, the $i$-th diagonal entry $m_i$ of $\V{M}$ is 
\begin{equation}
    m_i = \begin{cases}
        1, \quad \text{if $i$-th entry of \depthFront in \domainD} \\
        0, \quad \text{otherwise}
    \end{cases},
\end{equation}
while the $i$-th diagonal entry $s_i$ of $\V{S}$ is
\begin{equation}
    s_i = \begin{cases}
        1, \quad \text{if $i$-th entry of \depthFront in \silhouette}\\
        0, \quad \text{otherwise}
    \end{cases}.
\end{equation}
Stacking \depthFront and \depthBack as 
$\depthCombined = \left[\begin{smallmatrix}
        \depthFront \\
        \depthBack
    \end{smallmatrix}
    \right]$,
\cref{eq: dbni_opt_long} then reads
\begin{equation}
\begin{aligned}
    \mathcal{L}(\depthCombined)= (\V{A}\depthCombined - \V{b})^\top&\V{W}(\V{A}\depthCombined - \V{b}) + \\
    \lambda_\text{d}(\depthCombined - \depthCombinedPrior)^\top &\widetilde{\V{M}} (\depthCombined - \depthCombinedPrior) +
    \lambda_\text{s} \depthCombined^\top \widetilde{\V{S}} \depthCombined,
\end{aligned}
\label{eq: dbni_opt_short}
\end{equation}
where
% \begin{equation}
    \begin{gather*}
      \V{A} = \left[\begin{matrix}  
      \V{A}_{\text{F}} &  \\
       & \V{A}_{\text{B}}
      \end{matrix}\right], 
      \quad
      \V{b} = \left[\begin{matrix}
          \V{b}_{\text{F}} \\
          \V{b}_{\text{B}}
      \end{matrix} \right],
      \quad
      \V{W} = \left[\begin{matrix}  
      \V{W}_{\text{F}} &  \\
       & \V{W}_{\text{B}}
      \end{matrix}\right], \\
      \V{z} = \left[ \begin{matrix}
          \depthFrontPrior \\
          \depthBackPrior
      \end{matrix} \right] ,
      \quad
      \widetilde{\V{M}} = \left[\begin{matrix}  
      \V{M} &  \\
       & \V{M}
      \end{matrix}\right],
      \quad
      \widetilde{\V{S}} = \left[\begin{matrix}  
      \V{S} & -\V{S} \\
      -\V{S} & \V{S}
      \end{matrix}\right].
\end{gather*}
To minimize \cref{eq: dbni_opt_short}, we perform an iterative optimization similar to \bni~\cite{xu2022bilateral}. 
At each iteration, we first fix the weights $\V{W}$ and jointly solve for the front and back depth \depthCombined, then compute the new weights from the updated depth.
When $\V{W}$ is fixed and treated as a constant matrix, solving for the depth becomes a convex least-squares problem.
The necessary condition for the global optimum is obtained by equating the gradient of \cref{eq: dbni_opt_short} to $\V{0}$:
\begin{equation}
    \begin{aligned}
        (\V{A}^\top\V{W}\V{A} + \lambda_{\text{d}}\widetilde{\V{M}} + \lambda_{\text{s}}\widetilde{\V{S}})\depthCombined =
        \V{A}^\top\V{W}\V{b} + \lambda_{\text{d}}\widetilde{\V{M}} \depthCombinedPrior.
    \end{aligned}
\label{eq: dbni_linear_system}
\end{equation}
\Cref{eq: dbni_linear_system} is a large-scale sparse linear system with a symmetric positive definite coefficient matrix.
We solve \cref{eq: dbni_linear_system} using a CUDA-accelerated sparse conjugate gradient solver with a Jacobi preconditioner~\footnote{\url{https://docs.cupy.dev/en/stable/reference/generated/cupyx.scipy.sparse.linalg.cg.html}}.

\bigskip
\qheading{Hyper-parameters}
\dbni has three hyper-parameters: $\lambda_\text{d}$, $\lambda_\text{s}$, and $k$. $\lambda_\text{d}$ and $\lambda_\text{s}$ are used in the objective function~\cref{eq.dbni_functional}, which control the influence of coarse depth prior term ~\cref{eq.dbni_depth} and silhouette consistency term ~\cref{eq.dbni_boundary} separately. $k$ is used in the original \bni~\cite{xu2022bilateral} to control the surface stiffness (See Sup.Mat-A in \bni~\cite{xu2022bilateral} for more explanation of $k$). Empirically, we set $\lambda_\text{d} = 1e^{-4}$, $\lambda_\text{s} = 1e^{-6}$, and $k = 2$.

\bigskip
\qheading{Discussion of hyper-paramters}
\Cref{fig: dbni-K} shows the \dbni integration results under different values of $k$. It can be seen that a small $k$ leads to tougher \dbni surfaces where discontinuities are not accurately recovered, while a large $k$ softens the surface, and redundant discontinuities and noisy artifacts are introduced. \Cref{fig: dbni-depth} shows the effects of $\lambda_\text{d}$, which controls how much \dbni surfaces agree on the \smplx mesh. Small $\lambda_\text{d}$ causes misalignment between the \dbni surface and the \smplx mesh, which will produce stitching artifacts. While an excessively large $\lambda_\text{d}$ enforces \dbni to rely over heavily on \smplx, thus smoothing out the high-frequency details obtained from normals. \Cref{fig: dbni-bc} justifies the necessity of the silhouette consistency term. Without this term, the front and back \dbni surfaces intersect each other around the silhouettes, which will cause ``blobby'' artifacts after screened Poisson reconstruction~\cite{kazhdan2013screened}.

\begin{figure}[h]
    \centering
    % \vspace{-0.2in}
    \includegraphics[trim=000mm 000mm 000mm 000mm, clip=true, width=1.00\linewidth]{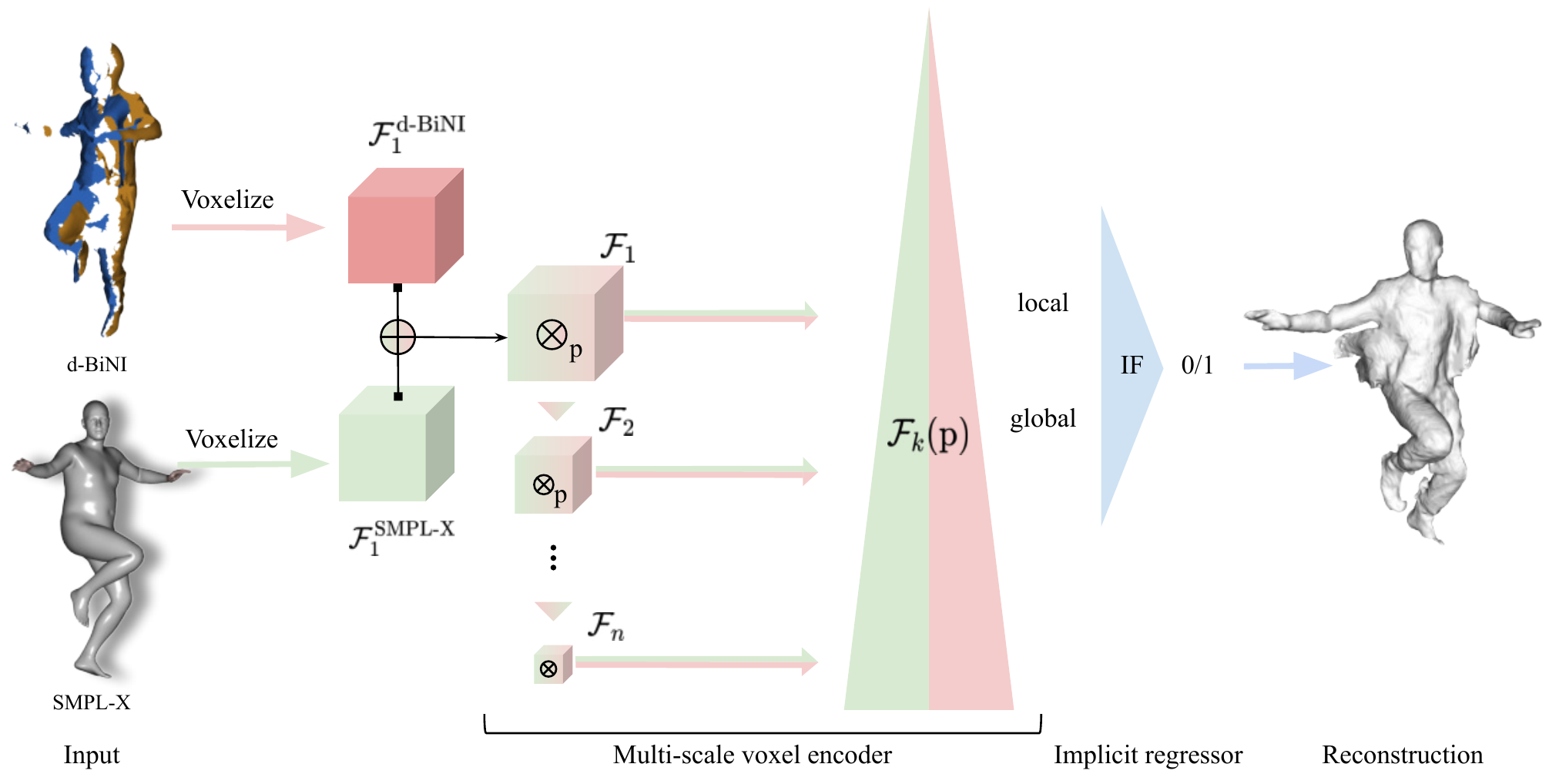}
    \caption{\qheading{Overview of \ifnetplus}}
    \vspace{-0.1in}
    \label{fig: ifnet+}
\end{figure}

\subsection{\ifnetplus}
\label{sec: suppl-ifnetplus}
\qheading{Network structure}
As~\cref{fig: ifnet+} shows, similar to \ifnet~\cite{chibane20ifnet}, \ifnetplus applies multi-scale voxel 3D CNN encoding on voxelized \dbni and the \smplx surface, namely $\mathcal{F}_\text{1}^\text{\dbni}$ and $\mathcal{F}_\text{1}^\text{\smplx}$, generating multi-scale deep feature grids \change{to account for both local and global information}, $\mathcal{F}_\text{1}$, $\mathcal{F}_\text{2}$, \dots, $\mathcal{F}_n$, $\mathcal{F}_k \in \mathbb{R}^{K \times K \times K \times C_k}, n=6$. 
These deep features are with decreasing resolution $K = \frac{N}{2^\text{k}-1}, N = 256$ and variable dimension channels $C = \{32, 32, 64, 128, 128, 128\}$. 
% Following~Li~\etal~\cite{Li2022SHARP}, the positional embedding ($\text{N}\_\text{freqs}=6$) of query points is concatenated with multi-scale deep features to account for high-frequency details. 
All these features are then fed into an implicit function regressor, parameterized by a Multi-Layer Perceptron (MLP), to predict the occupancy value of point $\text{P}$. This MLP regressor is trained with BCE loss.

\bigskip
\qheading{Training setting}
\ifnet and \ifnetplus share the same training setting. The voxelization resolution for both \smplx and \dbni surfaces is $256^3$. We use RMSprop as an optimizer, with a learning rate $1e^{-4}$, and weight decay by a factor of 0.1 every 10 epochs. These networks are trained on an NVIDIA A100 for 20 epochs with a batch size of 48. Following \icon~\cite{xiu2022icon}, we sampled 10000 points with the mixture of cube-uniform sampling and surface-around sampling, with standard deviation of 5cm. 

\bigskip
\qheading{Dataset details}
We augment \thuman~\cite{tao2021function4d} by (1) rotating the scans every 10 degrees around the yaw axis, to generate $525 \times 36 = 18900$ samples in total, and (2) randomly selecting a rectangle region from the \dbni depth maps, and erasing its pixels~\cite{zhong2020random}. In particular, the erasing operation is being performed with $p=0.8$ probability, the range of aspect ratio of erased area is between 0.3 and 3.3, and its range of proportion are $\{0.01, 0.05, 0.2\}$.

\camera{\bigskip
\qheading{Speed analysis of \modelname vs. \icon}
\dbni takes 6.2 secs (150 iters). For $\text{\modelname}_{\text{IF}}$, the \ifnetplus plus \mcubes takes 2.6 secs (for $256^3$ resolution), and the Poisson step takes 10.7 secs (level=10). 
For a single image, $\text{\modelname}_{\text{IF}}$ takes 112 secs, and $\text{\modelname}_{\text{EX}}$ takes 97 secs. 
\icon, which shares the same \smplx fitting (w/ landmarks), takes 78 secs, and w/ cloth-refinement (50 iters) it takes 115 secs.}

\section{Qualitative results}
\label{suppl: more-results}
\Cref{fig: SHHQ} shows examples on SHHQ~\cite{fu2022stylegan}. \Cref{fig: reb-pamir} shows \pamir's results on the same photos in~\cref{fig: qualitative figure}. \Cref{fig: result-pose,fig: result-cloth,fig: result-fashion} show more comparisons used in our perceptual study, containing the results on \itw images with challenging poses, loose clothing, and standard fashion poses, respectively.
For each image, we display the results obtained by \modelname, \pamir~\cite{zheng2020pamir}, \icon~\cite{xiu2022icon}, and \pifuhd~\cite{saito2020pifuhd}. 
% from the top to the bottom row.
In each row, we show normal maps rendered in $\{0^\circ, 90^\circ, 180^\circ, 270^\circ\}$ views. The \video shows more reconstructions with a rotating virtual camera.

\begin{figure*}[h]
    \includegraphics[trim=000mm 000mm 000mm 000mm, clip=true, width=1.0\linewidth]{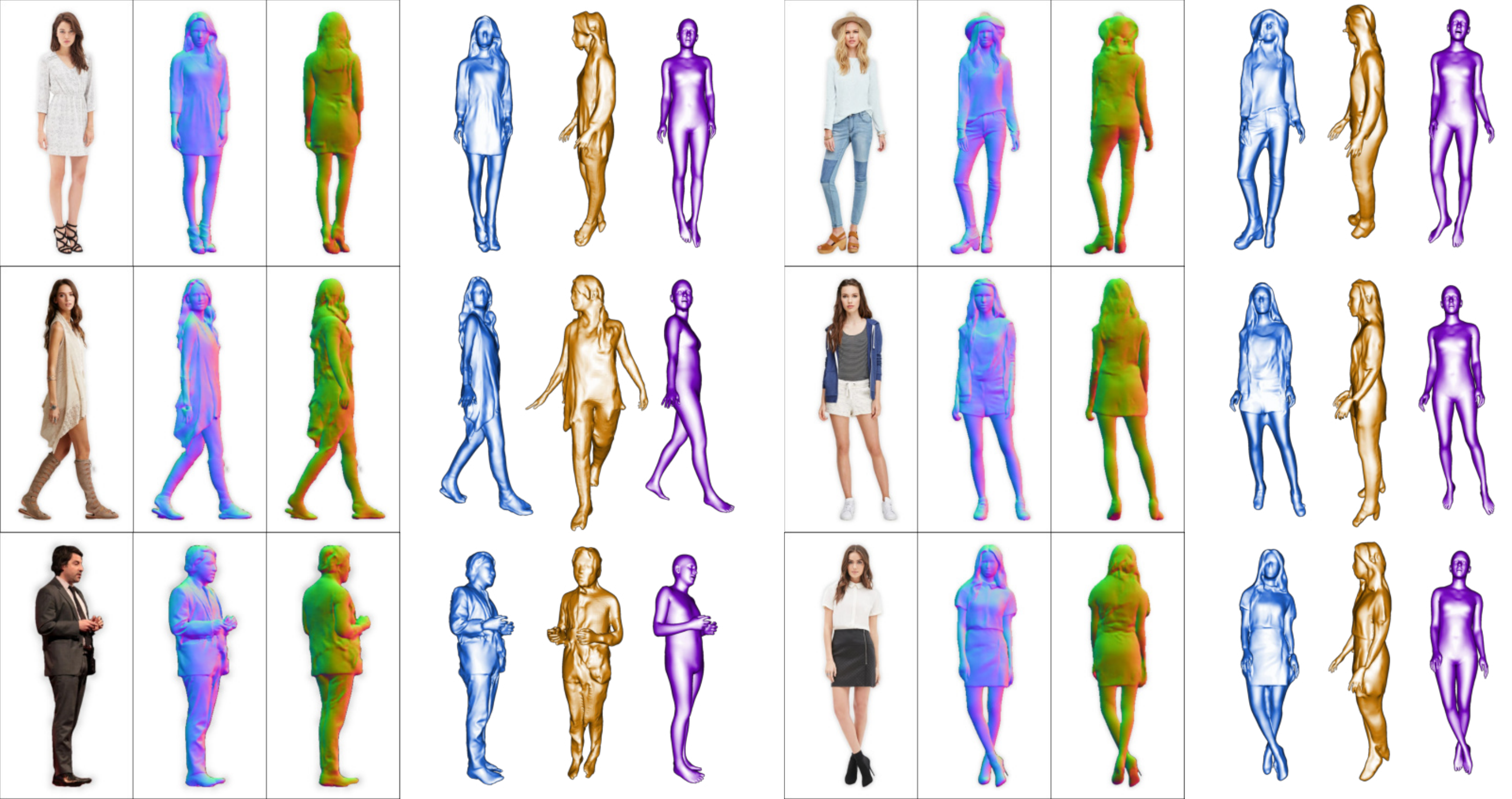}
    \caption{
                \cheading{SHHQ \threeD reconstruction} % augmentation
                For each image we show a 
                \textbf{\textcolor{frontcolor}{front}} and 
                \textbf{\textcolor{backcolor}{side}}  view of \modelname's reconstruction and a 
                \textbf{\textcolor{PurpleColor}{\smplx}} fit.
    }
    \label{fig: SHHQ}
    % \vspace{2em}
\end{figure*}

\begin{figure*}[h]
    \includegraphics[trim=000mm 000mm 000mm 000mm, clip=true, width=1.0\linewidth]{photos/OOD-outfits.pdf}
    \vspace{0.5 em}
    \includegraphics[trim=000mm 000mm 000mm 000mm, clip=true, width=1.0\linewidth]{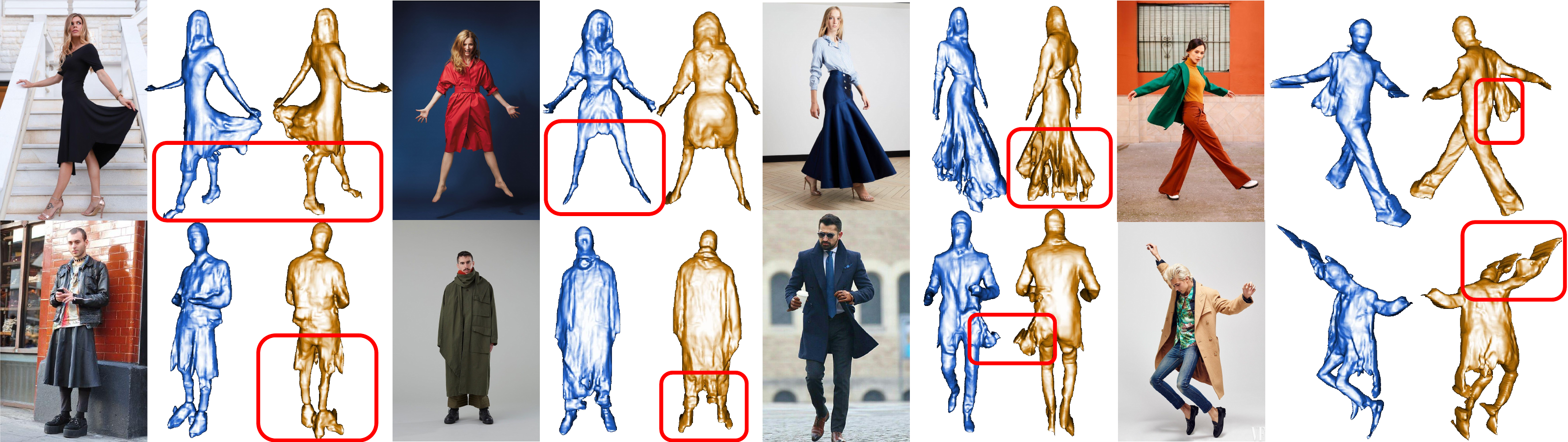}
    \vspace{-2.0 em}
    \caption{
                % \textbf{\pamir} on loose clothing; \faSearch~\textbf{Zoom in} to see details.
                \textbf{\modelname} (Top) vs. \textbf{\pamir} (Bottom) on loose clothes; \faSearch~\textbf{Zoom in} to see \textcolor{frontcolor}{front}/\textcolor{backcolor}{back} \threeD details. 
    }
    \vspace{-1.0 em}
    \label{fig: reb-pamir}
\end{figure*}

\clearpage

\begin{figure*}[t]
     \centerline{
    \includegraphics[trim=000mm 000mm 000mm 000mm, clip=true, width=0.95\linewidth]{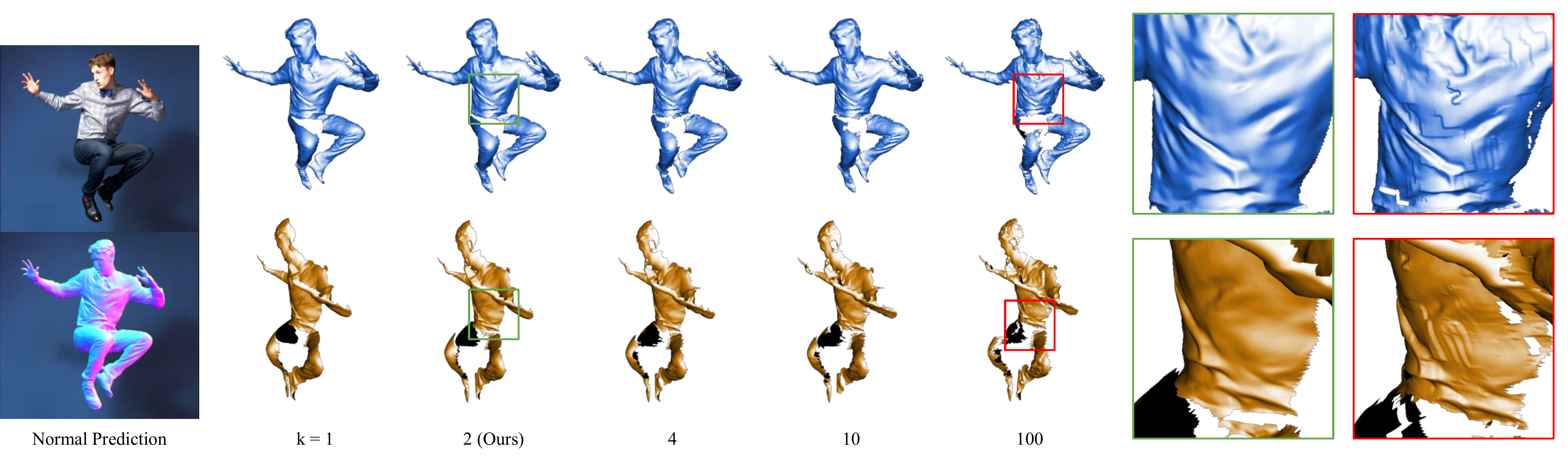}
    }
    \vspace{-0.1in}
    \caption{
            \qheading{The effects of the hyper-parameter $k$ on \dbni results}
            $k$ controls the stiffness of the target surface~\cite{xu2022bilateral}. 
            A smaller $k$ leads to smooth \dbni surfaces, while a large $k$ introduces unnecessary discontinuities and noise artifacts.
    }
    \vspace{-0.2in}
    \label{fig: dbni-K}
\end{figure*}
\begin{figure*}[t]
\centerline{
    \includegraphics[trim=000mm 000mm 000mm 000mm, clip=true, width=0.95\linewidth]{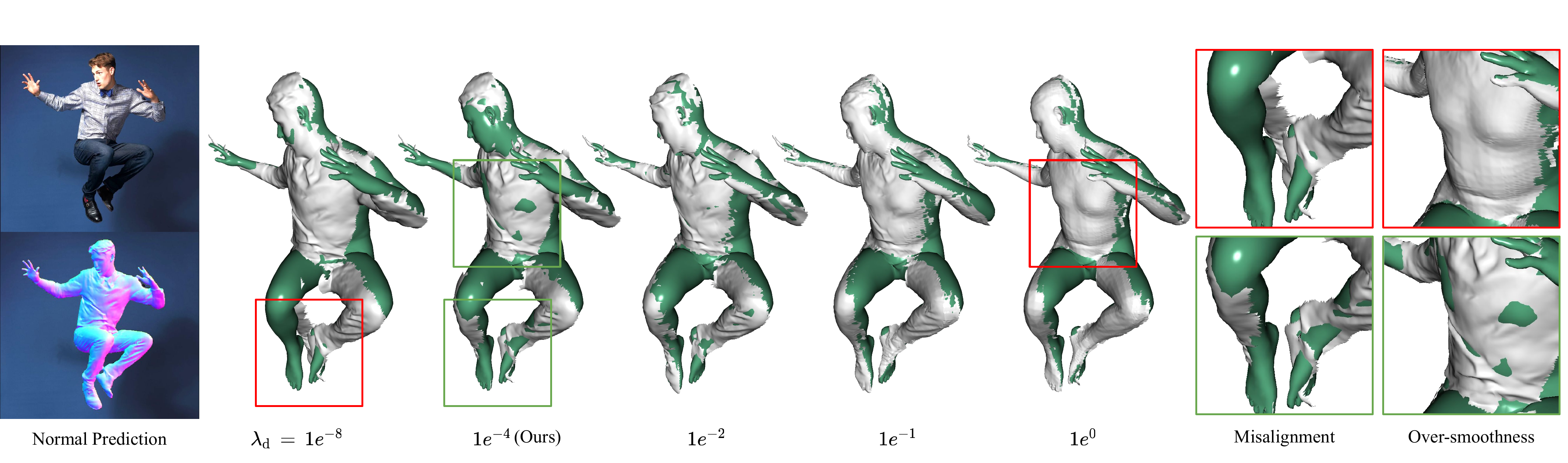}
    }
    \vspace{-0.1in}
    \caption{
                    \qheading{The effects of the hyperparameter $\lambda_\text{d}$ on \dbni results} 
                    $\lambda_\text{d}$ controls how much \dbni surfaces agree with the \smplx mesh. 
                    A small $\lambda_\text{d}$ causes a misalignment between the \dbni surface and the \smplx mesh, thus it produces stitching artifacts. 
                    An excessively large $\lambda_\text{d}$ enforces \dbni to rely too heavily on \smplx, thus it smooths out the high-frequency details obtained from normals. 
    }
    \label{fig: dbni-depth}
    \vspace{-0.1in}
\end{figure*}
\begin{figure*}[th!]
\centerline{
    \includegraphics[trim=000mm 000mm 000mm 000mm, clip=true, width=0.95\linewidth]{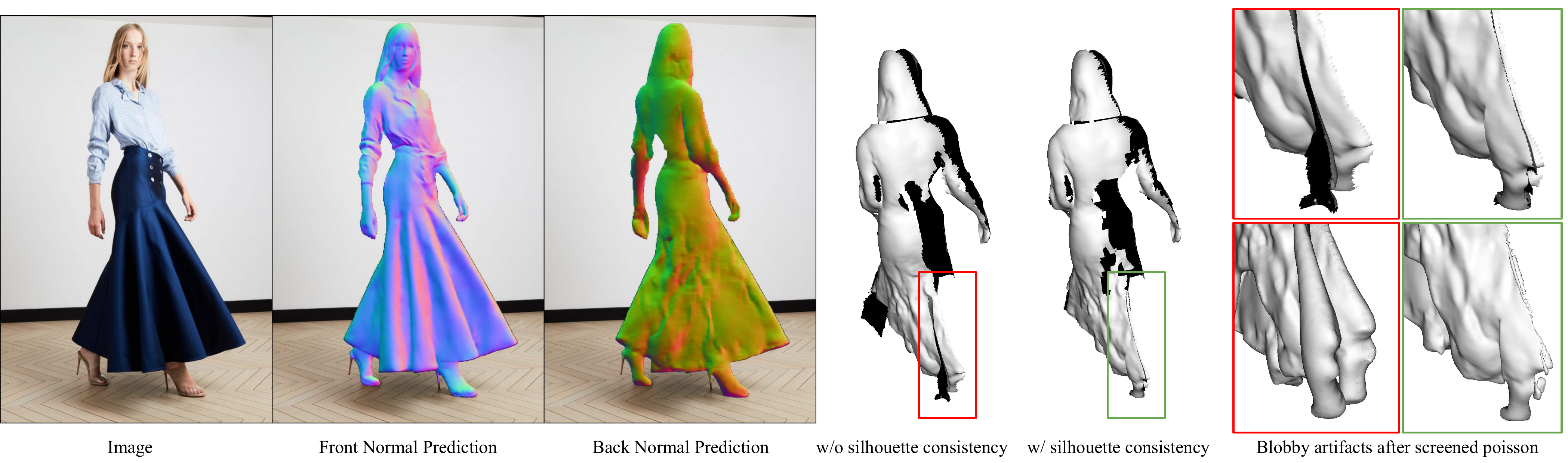}
    }
    \vspace{-0.1in}
    \caption{
                    \qheading{Necessity of silhouette consistency} 
                    This term can be regarded as the mediator between front and back \dbni surfaces, preventing these surfaces from intersecting. 
                    Such intersection causes blobby artifacts after screened Poisson reconstruction~\cite{kazhdan2013screened}.
    }
    \label{fig: dbni-bc}
\end{figure*}

\begin{figure*}[t]
\vspace{-0.7in}
\centerline{
    \includegraphics[trim=000mm 000mm 000mm 000mm, clip=true, width=0.90\linewidth]{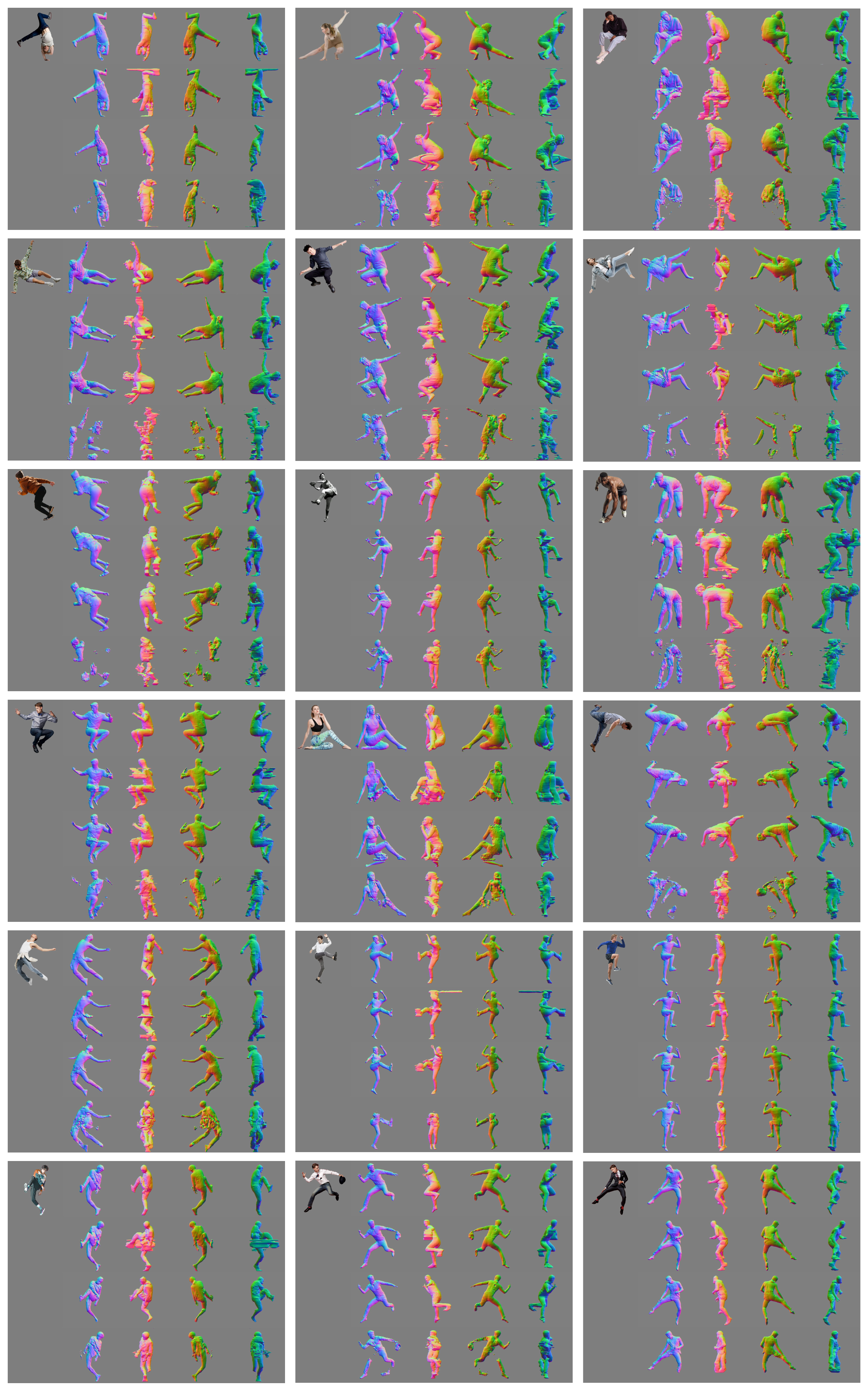}
    }
    \vspace{-0.1in}
    \caption{
                \qheading{Results on \itw images with challenging poses} 
                For each example the format is as follows:  
                \textbf{\mbox{Top $\rightarrow$ bottom:}} \modelname, \pamir~\cite{zheng2020pamir}, \icon~\cite{xiu2022icon}, and \pifuhd~\cite{saito2020pifuhd}. 
                \textbf{\mbox{Left $\rightarrow$ right:}} Virtual camera rotated by $\{0^\circ, 90^\circ, 180^\circ, 270^\circ\}$. 
                \faSearch~\textbf{Zoom in} to see \threeD details.
    }
    \label{fig: result-pose}
\end{figure*}

\begin{figure*}
\vspace{-0.7in}
\centerline{
    \includegraphics[trim=000mm 000mm 000mm 000mm, clip=true, width=0.90\linewidth]{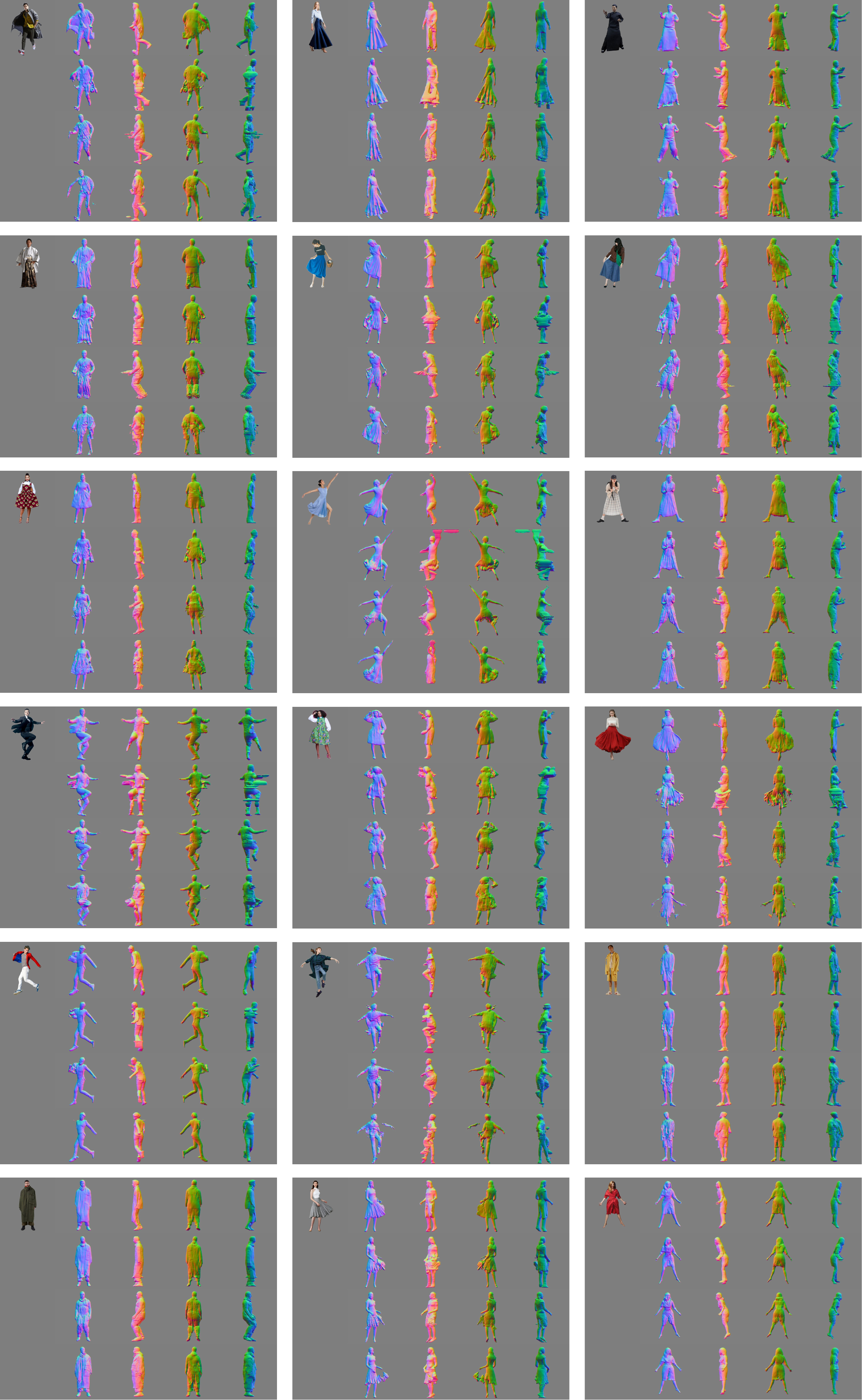}
    }
    \vspace{-0.1in}
    \caption{
                \qheading{Results on \itw images with loose clothing} 
                For each example the format is as follows:  
                \textbf{\mbox{Top $\rightarrow$ bottom:}} \modelname, \pamir~\cite{zheng2020pamir}, \icon~\cite{xiu2022icon}, and \pifuhd~\cite{saito2020pifuhd}. 
                \textbf{\mbox{Left $\rightarrow$ right:}} Virtual camera rotated by $\{0^\circ, 90^\circ, 180^\circ, 270^\circ\}$. 
                \faSearch~\textbf{Zoom in} to see \threeD details.
    }
    \label{fig: result-cloth}
\end{figure*}
\begin{figure*}
\vspace{-0.7in}
\centerline{
    \includegraphics[trim=000mm 000mm 000mm 000mm, clip=true, width=0.90\linewidth]{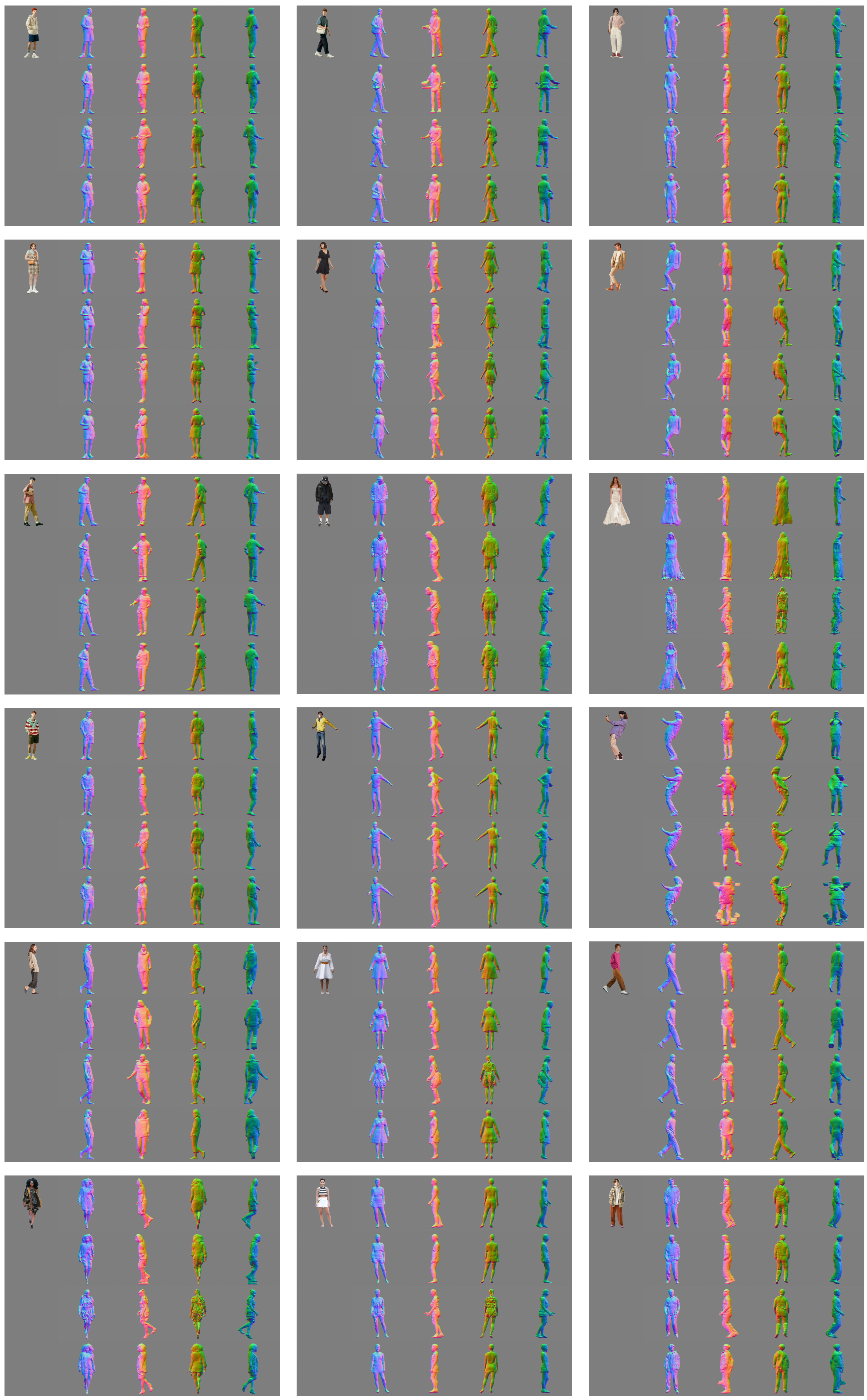}
    }
    \vspace{-0.1in}
    \caption{
                \qheading{Results on \itw fashion images} 
                For each example the format is as follows:  
                \textbf{\mbox{Top $\rightarrow$ bottom:}} \modelname, \pamir~\cite{zheng2020pamir}, \icon~\cite{xiu2022icon}, and \pifuhd~\cite{saito2020pifuhd}. 
                \textbf{\mbox{Left $\rightarrow$ right:}} Virtual camera rotated by $\{0^\circ, 90^\circ, 180^\circ, 270^\circ\}$. 
                \faSearch~\textbf{Zoom in} to see \threeD details.
    }
    \label{fig: result-fashion}
\end{figure*}

\end{appendices}

\clearpage
{\small
\balance
\bibliographystyle{config/ieee_fullname}
\bibliography{config/BIB}
}

\end{document}